\documentclass[final,12pt]{colt2023} 

\title[Online Subset Selection using $\alpha$-\textsc{Core} with no Augmented Regret]{Online Subset Selection using $\alpha$-\textsc{Core} with no Augmented Regret}
\usepackage{times}
\usepackage{amsmath}
\usepackage{adjustbox}
\usepackage{autonum}
\usepackage{dsfont}
\usepackage{booktabs}
\usepackage{amssymb}
\usepackage{algorithm}
\usepackage{algorithmic}
\usepackage{tikz}
\usepackage{graphics}
\usepackage{bm}
\usepackage{nicefrac}

\newcommand{\edit}[1]{{\color{black} #1}}

\newcommand{\onscore}{\textsc{SCore }}
\newcommand{\core}[2]{\mathrm{\textsc{Core}}_{#1}\left(#2\right)}
\newcommand{\norm}[1]{\left\lVert#1\right\rVert}

\coltauthor{%
 \Name{Sourav Sahoo} \Email{sourav22899@alumni.iitm.ac.in}\\
 \addr Dept. of Electrical Engineering \\
 \addr Indian Institute of Technology Madras
 \AND
 \Name{Siddhant Chaudhary} \Email{chaudhary@cmi.ac.in}\\
 \addr Chennai Mathematical Institute, Chennai, India
 \AND 
 \Name{Samrat Mukhopadhyay} \Email{samrat@iitism.ac.in}\\
 \addr Dept. of Electronics Engineering \\
 \addr Indian Institute of Technology (ISM) Dhanbad
 \AND
 \Name{Abhishek Sinha} \Email{abhishek.sinha@tifr.res.in}\\
 \addr School of Technology and Computer Science \\
 \addr Tata Institute of Fundamental Research, Mumbai, India
}

\begin{document}

\maketitle

\begin{abstract} \label{abstract}
We revisit the classic problem of optimal subset selection in the online learning set-up. Assume that the set $[N]$ consists of $N$ distinct elements. On the $t$\textsuperscript{th} round, an adversary chooses a monotone reward function $f_t: 2^{[N]} \to \mathbb{R}_+$ that assigns a non-negative reward to each subset of $[N].$ An online policy selects (perhaps randomly) a subset $S_t \subseteq [N]$ consisting of $k$ elements before the reward function $f_t$ for the $t$\textsuperscript{th} round is revealed to the learner. As a consequence of its choice, the policy receives a reward of $f_t(S_t)$ on the $t$\textsuperscript{th} round. Our goal is to design an online sequential subset selection policy to maximize the expected cumulative reward accumulated over a time horizon. In this connection, we propose an online learning policy called \onscore (\textbf{S}ubset Selection with \textbf{Core}) that solves the  problem for a large class of reward functions. The proposed \onscore policy is based on a new polyhedral characterization of the reward functions called \emph{$\alpha$-\textsc{Core}} - a generalization of Core from the cooperative game theory literature. We establish a learning guarantee for the \onscore policy in terms of a new performance metric called $\alpha$-\emph{augmented regret}. In this new metric, the performance of the online policy is compared with an unrestricted offline benchmark that can select all $N$ elements at every round. We show that a large class of reward functions, including submodular, can be efficiently optimized with the \onscore policy. We also extend the proposed policy to the optimistic learning set-up where the learner has access to additional untrusted hints regarding the reward functions. Finally, we conclude the paper with a list of open problems. 
\end{abstract}

\section{Introduction} \label{intro}
\emph{Subset Selection} problems refer to a broad family of computational problems where the overall goal is to efficiently select a small number of elements from a larger ground set to maximize a monotone set function. Optimal subset selection is fundamental to many tasks in AI and ML, including feature selection \citep{khanna2017scalable}, coreset selection \citep{borsos2020coresets}, sparse regression \citep{kale2017adaptive}, and viral marketing in social networks \citep{kempe2003maximizing}. 
Because of its practical utility and theoretical significance, both the offline and online variants of the subset selection problem have been extensively studied in the literature. We refer the readers to Section \ref{related-works} for a brief survey of the related works.
 In the online setting, a particularly important variant of the problem is the classic \emph{Prediction with Expert advice}, where the learner randomly selects only one  expert from the set of $N$ experts at each round to optimize a linear reward function \citep{cesa2006prediction}. 
In this paper, we consider the problem of sequentially selecting a subset of $k$ elements from a universe of $N$ elements to maximize the sum of a sequence of non-negative monotone reward functions. The reward functions, which could be adversarially chosen, are revealed to the learner at the end of each round in an online fashion. 

In case of submodular reward functions, 
\citet{streeter2007online} designed the first online subset selection policy with a sublinear regret. Their proposed policy greedily runs $k$ instances of an expert algorithm at each round. Several follow-up works have extended the scope and improved the complexity of their original policy. See Table \ref{submodular-comp-table} for a comparison of our proposed \onscore policy with other existing algorithms.


\subsection*{Our Contributions}
\edit{In contrast with the vast literature on the online subset selection problem, we take a fresh look at this problem through the lens of a natural \emph{polyhedral} characterization of the reward functions. Analogous to the classic Zinkevich step \citep{zinkevich2003online} that reduces the general OCO problem to online linear optimization, the polyhedral characterization essentially reduces the combinatorial subset selection problem to a simpler linear reward variant. This approach yields simpler and more efficient algorithms with new regret bounds for a large class of reward functions.} In particular, we make the following main contributions:
\begin{enumerate}
	\item In Section \ref{characterization}, we introduce the notion of $\alpha$-\textsc{Core} polytope, which generalizes the notion of Core from game theory and Base polytope from polyhedral combinatorics. We present a dual characterization of $\alpha$-\textsc{Core} in Theorem \ref{ex-bs}, which is then used to derive simple sufficient conditions for the existence of a non-empty $\alpha$-\textsc{Core} (see Proposition \ref{dictator-prop} in the Appendix). 
	\item In Section \ref{regret-bd}, we propose a new efficient online subset selection policy called \textsc{SCore}.
	Theorem \ref{main-thm} gives a \emph{dimension-free} learning guarantee for the \onscore policy for all $\alpha$-admissible reward functions. 
	The flexibility of our framework allows us to utilize results from the cooperative game theory literature in the online learning setup (see Section \ref{non-submod1} in the Appendix). Section \ref{admissible-rewards} lists several examples to show that the proposed \onscore policy can efficiently maximize a broad class of reward functions with significant practical utility.
	
	\item We introduce a new learning metric called \emph{augmented regret}. In augmented regret, the offline benchmark is taken to be a fraction of the maximum cumulative reward that could have been obtained by selecting \emph{all} elements at each round. Unlike static or dynamic regret, the augmented regret metric does not depend on any specific \emph{ad hoc} choice of stationary or semi-stationary benchmarks and might be of independent interest. \edit{Furthermore, we show that our augmented regret bound improves upon the standard $(1-e^{-1})$-static regret bound for maximizing monotone submodular functions when $\nicefrac{k}{N} \geq 1-e^{-1}.$} The notion of augmented regret is inspired by \emph{resource augmentation} that has been used in the competitive analysis of online algorithms \citep{sleator1985amortized}.
	\edit{\item Finally, we extend our online policy to the optimistic learning set-up when the reward functions are submodular and some modular hints (\emph{a.k.a.} advice) are available to the learner before it makes the decision. In Theorem \ref{oftrlBound}, we present an optimistic augmented regret bound that scales with the \emph{quality} of the hints. To the best of our knowledge, \onscore is the first online learning policy for efficiently maximizing submodular functions while exploiting hints.}
	
	

\item Although in this paper, we work exclusively in the full-information setting, an extension to incomplete information set-up has been given in Appendix \ref{priced}. In this \emph{priced feedback} model,  we assume that the learner, if it chooses so, pays a fixed cost at each round to know the reward function. We show that the \onscore policy achieves $O(T^{2/3})$-augmented regret in this setting. 
\end{enumerate}
\begin{table}
  \caption{Comparison among different policies for the \textsc{Online Subset Selection} problem*}
  \label{submodular-comp-table}
  \centering
  \begin{adjustbox}{width=1\textwidth}
\begin{tabular}{cccccc}
    \toprule
    Policies     & Scope & Approx. & Regret   & Complexity & Learning with Hints \\
    \midrule
    \cite{streeter2007online} & submodular & $1-e^{-1}$ & $O(\sqrt{kT\ln N})$ & $O(Nk)$ & NA\\
    \cite{harvey2020improved} & submodular & $1-e^{-1}$ & $O\left(\sqrt{kT\ln \frac{N}{k}}\right)$& $O\left(\frac{N^4}{\epsilon^3}\ln\frac{N^3T}{\epsilon}\right) +O(\frac{N^3}{\epsilon})\cdot \text{SFM}$ & NA\\
     \cite{kale2017adaptive} & weakly supermodular& $1-e^{-1}$ & $\tilde{O}(k\sqrt{T\ln N})$& $O(Nk)$ & NA\\
   \textbf{\onscore} & $\alpha$-admissible & $\alpha^{-1}$ & $O\left(\sqrt{kT \ln \frac{N}{k}}\right)$\textsuperscript{**}& $\tilde{O}(N)+\text{CFS}(\alpha)$ & \textbf{Yes} \\
    \bottomrule
  \end{tabular}
  \end{adjustbox}
  {
  \raggedright \small{
*In the above table $\epsilon$ denotes the accuracy parameter, SFM denotes the time complexity for minimizing a submodular function ~\citep{harvey2020improved}, and CFS($\alpha$) denotes the time complexity for finding an $\alpha$-admissible vector for a reward function. The approximation ratio is stated with reference to the corresponding minimax regret lower bound. \\
\textsuperscript{**} Augmented regret bound. See Eqn.\ \eqref{regret-guarantee1} for the definition of augmented regret. \vspace{2pt} \hrule\hrule}\par}
\end{table}
\section{Preliminaries} \label{sum-reward-section}
We begin with the simpler \textsc{Linear-reward} version of the \textsc{Subset Selection} problem considered earlier by \citet{aistats22}. In this problem, the adversary chooses a linear (\emph{i.e.,} modular) function $f_t(S) := \sum_{i \in S} g_{t,i}, \forall S \subseteq [N]$ at every round $t\geq 1$. The learner randomly selects a $k$-set $S$ with probability $p_t(S), S \subseteq [N]$ and receives an expected reward of $\sum_{S} p_t(S)f_t(S)$ on the $t$\textsuperscript{th} round.  The reward function $f_t$ (equivalently, the coefficient vector $\bm{g}_t$) is revealed to the learner at the end of round $t.$
The learner's objective is to maximize its cumulative reward accumulated over a horizon of length $T$. \citet{aistats22} made two key observations about this problem:
\begin{enumerate}
\item Because of the \emph{linearity} of the reward function $f_t$, the expected reward at any round depends only on the marginal inclusion probabilities of the elements, \emph{i.e.,}
\begin{eqnarray} \label{linearity}
	\mathbb{E}[f_t(S)] = \sum_{i=1}^N p_t(i) g_{t,i},
\end{eqnarray} 
where $p_t(i)\equiv \sum_{S \ni i} p_t(S)$ is the marginal inclusion probability of the $i$\textsuperscript{th} element in $S_t$. 
\item The following two conditions \eqref{necsuf1} and \eqref{necsuf2} are necessary and sufficient for a vector $\bm{q}$ to correspond to the marginal inclusion probabilities induced by some sampling scheme that samples $k$ elements from $N$ elements \emph{without replacement}: 
\begin{eqnarray} 
	&&\sum_{i=1}^N q_i = k, \label{necsuf1}\\
	&&0\leq q_i \leq 1,~ \forall i \in [N]\label{necsuf2} .
\end{eqnarray}
In other words, conditions \eqref{necsuf1} and \eqref{necsuf2} are equivalent to the existence of a sampling distribution $\{p(S), S \subseteq [N]\}$, such that 
\begin{eqnarray*}
	\sum_{S: |S|=k} p(S)=1,~~~ \textrm{and}~~~ 
	q_i = \sum_{S \ni i} p(S), ~\forall i \in [N].
\end{eqnarray*}
\end{enumerate}
The necessity of condition \ref{necsuf2} is obvious. Furthermore, since the online policy samples exactly $k$ elements without replacement, condition \ref{necsuf1} follows by enforcing this condition in expectation.   
Let $\Delta^N_k$ denote the set of all vectors satisfying requirements \eqref{necsuf1} and \eqref{necsuf2}. \edit{In the polyhedral combinatorics literature, the set $\Delta_k^N$ is known as the $(k,N)$-\emph{hypersimplex} \citep{rispoli2008graph}.} Given any vector $\bm{q} \in \Delta^N_k,$ one can use a systematic sampling scheme, such as Madow's sampling, to efficiently sample a $k$-set in $O(N)$ time such that the $i$\textsuperscript{th} element is included in the sampled set exactly with probability $q_i, \forall i \in [N].$
For completeness, we outline Madow's sampling scheme in Appendix \ref{madow-section}.

  Since the set of feasible inclusion probabilities $\Delta^N_k,$ given by \eqref{necsuf1} and \eqref{necsuf2}, is convex, one can use the standard Online Convex Optimization framework for sequentially learning a suitable marginal inclusion probability $\bm{p}_{t} \in \Delta^N_k$ at every round \citep{hazan2019introduction}. The paper by \cite{aistats22} uses the \emph{Follow-the-Regularized-Leader} (FTRL) framework with the entropic regularizer, that computes the marginal inclusion probabilities by solving the following optimization problem:
  \begin{eqnarray} \label{ftrl-opt}
\bm{p}_t = \arg\max_{\bm{p} \in \Delta^N_k} \left[\left(\sum_{s=1}^{t-1} \bm{g}_s\right)^\top \bm{p} - \frac{1}{\eta}\sum_{i=1}^N p_i \ln p_i,\right].
\end{eqnarray}
  The authors show that the above optimization problem can be efficiently solved in $\tilde{O}(N)$ time per round.  
One can then use Madow's sampling scheme to sample $k$ elements without replacement. 
  Their overall online subset selection policy is summarized below in Algorithm \ref{linear-reward-algo}.   \begin{algorithm}
\caption{Online Subset Selection Policy for Linear Rewards}
\label{linear-reward-algo}
\begin{algorithmic}[1]
\STATE \textbf{Input:} Sequence of linear reward functions $\{f_t\}_{t \geq 1}$ with coefficients $\{\bm{g}_t\}_{t\geq 1}, ||\bm{g}_t||_2 \leq G, \forall t.$
\STATE \textbf{Output:} Sampled $k$-sets $\{S_t\}_{t\geq 1}.$
\FOR {every round $t$:}
\STATE Efficiently compute the marginal inclusion probability vector $\bm{p}_t$ in $\tilde{O}(N)$ time using \textsc{FTRL} (Eqn.\ \eqref{ftrl-opt}) while taking $\eta = \sqrt{\frac{k \ln(N/k)}{2G^2T}}.$
\STATE Efficiently sample a $k$-set $S_t$ according to the inclusion probability vector $\bm{p}_t$ using Madow's sampling scheme (Algorithm \ref{uneq}).
\STATE Receive the coefficient vector $\bm{g}_t$ of the linear reward function $f_t.$ 
\ENDFOR
\end{algorithmic}
\end{algorithm}

The following result gives an upper bound to the static regret achieved by Algorithm \ref{linear-reward-algo}.

  \begin{theorem} [Regret bound for \textsc{Linear-rewards} \citep{aistats22}] \label{lin-reward}
  	The online policy in Algorithm \ref{linear-reward-algo} yields the following standard regret bound for the \textsc{Linear-Reward} problem: 
  	\begin{eqnarray} \label{reg-bd-lin}
  		\max_{\bm{p}^* \in \Delta^N_k} \sum_{t \leq T} \langle \bm{g}_t, \bm{p}^* \rangle - \sum_{t \leq T} \langle \bm{g}_t, \bm{p}_t \rangle \leq 2G\sqrt{2kT\ln \frac{N}{k}},
  	\end{eqnarray}
  		where the variable $G$ upper-bounds the $\ell_2$-norm of each of the coefficient vectors $\{\bm{g}_t\}_{t=1}^T.$  
  \end{theorem}
  
  \paragraph{Remarks:} Instead of using the entropic regularizer in \eqref{ftrl-opt}, one can also use other strongly convex regularizers by striking a trade-off between the tightness of the regret bound and run-time complexity. For example, the logarithmic factor from the regret bound \eqref{reg-bd-lin} can be removed by using the quadratic regularizer at the expense of a slightly more computationally expensive projection step. Please refer to Section \ref{learningwithhints}, which uses adaptive Euclidean regularizers and avoids the projection steps altogether with Frank-Wolfe iterations. Our proposed \onscore policy uses  an idea similar to Algorithm \ref{linear-reward-algo} to maximize a broad class of non-linear reward functions, as described next. 
\section{Admissible Rewards} \label{characterization}
In this section, we define and characterize the class of admissible reward functions that can be learned by our proposed online policy \textsc{SCore}, described in the following section. In Appendix \ref{reward-example}, we give several examples to show that the class of admissible rewards is sufficiently rich and includes a number of standard reward functions encountered in practice. 

Recall that a function $f: 2^{[N]} \to \mathbb{R}$ is monotone non-decreasing when $f(S)\le f(T), \forall S\subseteq T$. Furthermore, we assume that all reward functions are normalized, \emph{i.e.,} $f_t(\emptyset) = 0, \forall t$. 
\begin{definition}[$\alpha$-admissible rewards] \label{admissible-rew}
	A non-negative reward function $f: 2^{[N]} \to \mathbb{R}_{+} $ is called $\alpha$-admissible ($\alpha \geq 1$) iff 
		there exists a set of real numbers $g_1, g_2, \ldots, g_N$ such that 
		\begin{eqnarray} \label{core-def1}
			 \sum_{i \in S} g_i &\leq& \alpha f(S), ~~ \forall S \subseteq [N],\\
			 \sum_{i=1}^N g_i &=&  f([N]). \label{core-def2}
		\end{eqnarray}

The class of all $\alpha$-admissible reward functions is denoted by $\mathcal{F}_\alpha.$ The set of all vectors $\bm{g} \in \mathbb{R}^N$ satisfying \eqref{core-def1} and \eqref{core-def2} is called the $\alpha$-core of the function $f$ and is denoted by $\core{\alpha}{f}$. For an $\alpha$-admissible function $f,$ any vector $\bm{g} \in \core{\alpha}{f}$ is called $\alpha$-admissible.	
\end{definition}
It immediately follows from Definition \ref{admissible-rew} that the class $\{\mathcal{F}_\alpha\}_{\alpha \geq 1}$ is monotone non-decreasing, \emph{i.e.,} $\mathcal{F}_{\alpha_2} \supseteq \mathcal{F}_{\alpha_1}, \forall \alpha_2 \geq \alpha_1 \geq 1.$ Linear (\emph{i.e.,} modular) functions are trivially $1$-admissible. Note that class $\mathcal{F}_1$ coincides with the set of all transferable-utility (TU) coalitional cost-sharing games $(N, f)$ with a non-empty \emph{Core} \citep{shapley1971cores, maschler2020game}\footnote{In the more common \emph{profit-sharing} paradigm considered in the game-theoretic literature, the direction of the inequalities in \eqref{core-def1} is reversed in the definition of Core.}. In the terminology of cooperative game theory, a $1$-admissible vector
is known as a \emph{coalitionally rational imputation} of the game $(N,f)$ \citep{maschler2020game}. \edit{In the combinatorial optimization literature, for a  submodular function $f$, the polytope $\textsc{Core}_1(f)$ is also known as the \emph{Base polytope}. }


The question of the existence of a non-empty core for different classes of games has been extensively investigated in the cooperative game theory literature. In general, determining the non-emptiness of the core of an arbitrary game 
is known to be \textbf{NP-Complete} \citep[Theorem 1]{conitzer2003complexity}. The celebrated Bondareva-Shapley theorem gives a necessary and sufficient condition for a TU coalitional game to have a non-empty core. 
In the following, we extend this result to the class of admissible rewards $\{\mathcal{F}_\alpha, \alpha \geq 1\}$. 
\begin{theorem}[Existence of a non-empty $\alpha$-\textsc{Core}] \label{ex-bs}
	A reward function $f: 2^{[N]}\to \mathbb{R}$ is $\alpha$-admissible iff for every function $\delta : 2^{[N]} \to [0,1] $ where $\sum_{S: i \in S} \delta(S) = 1, \forall i \in [N]$, we have
	\begin{equation}
		\alpha \sum_{S \subseteq [N]} \delta(S)f(S) \geq f([N]).
	\end{equation}
\end{theorem}
We prove Theorem \ref{ex-bs} in Appendix \ref{ex-bs-proof} by using the dual of the LP given in Definition \ref{admissible-rew}. 
\paragraph{Discussion:}
Theorem \ref{ex-bs} yields the following alternative interpretation for the $\alpha$-\textsc{Core}. Note that the condition $\sum_{S: i \in S} \delta(S) = 1, \forall i \in [N]$ is equivalent to requiring that the variables $\{\delta(S), S \subseteq [N]\}$ fractionally cover all elements of the set $[N]$. Now consider an instance of the \textsc{Weighted Set Cover} problem on the system $([N], 2^{[N]})$ where the cost for selecting a subset $S$ is defined to be $c(S)\equiv \nicefrac{f(S)}{f([N])}.$ If the minimum cost to fractionally cover all elements in $[N]$ is $\textsc{OPT}^*,$ then the smallest value of $\alpha$ for which $\core{\alpha}{f}$ is non-empty is given by $1/\textsc{OPT}^*$. 
Finally, the following structural result shows that the $\alpha$-\textsc{Core} of any reward function, taking values in the interval $[0,M],$ entirely lies within the Euclidean ball of radius $\alpha M \sqrt{2}$ centred at the origin. Surprisingly, the radius of the bounding ball is independent of the ambient dimension $N.$ As we will see later in Theorem \ref{main-thm}, this result is the key to our \emph{dimension-free} (up to a logarithmic factor) augmented regret bound \eqref{regret-guarantee1}.
\begin{proposition} \label{core-bd}
		Let $f$ be an $\alpha$-admissible reward function taking values in the interval $[0,M].$
		Then for any $\bm{g} \in \core{\alpha}{f},$ we have $||\bm{g}||_2 \leq \alpha M \sqrt{2}$.
\end{proposition} 
See Appendix \ref{core-bd-proof} for the proof of Proposition \ref{core-bd}. In the particular case of monotone $\rho$-submodular reward functions, Proposition \ref{beta-submod-norm} in Appendix \ref{core-bd-proof} gives a slightly tighter bound. 
\section{\onscore - an Efficient Online Subset Selection Policy}\label{regret-bd}
In this section, we propose \onscore - a new efficient online subset selection policy. 
On a high level, the \onscore policy uses an appropriately chosen linear reward function $\bm{g}_t$ as a proxy to the reward function $f_t.$ It then
uses Algorithm \ref{linear-reward-algo} for predicting an appropriate $k$-set. In particular, for $\alpha$-admissible reward functions, the \onscore policy chooses an $\alpha$-admissible vector $\bm{g}_t$ from the $\alpha$-\textsc{Core} of the reward function $f_t.$ The pseudocode for the \onscore policy is given below in Algorithm \ref{monotone-reward-algo}. The problem of efficiently computing $\alpha$-admissible vectors for various reward functions is discussed in Appendix \ref{reward-example}.

\begin{algorithm}
\caption{\onscore policy for $\alpha$-admissible rewards}
\label{monotone-reward-algo}
\begin{algorithmic}[1]
\STATE \textbf{Input:} Sequence of $\alpha$-admissible reward functions $\{f_t\}_{t\geq 1}, 0\leq f_t \leq M, \forall t.$
 \STATE \textbf{Output:} Sampled $k$-sets $\{S_t\}_{t\geq 1}.$
\FOR {every round $t$:}
\STATE Efficiently compute the marginal inclusion probability vector $\bm{p}_t$ in $\tilde{O}(N)$ time using \textsc{FTRL} (Eqn.\ \eqref{ftrl-opt}) while taking $\eta = \sqrt{\frac{k \ln(N/k)}{2G^2T}}$, where $G= \alpha M \sqrt{2}$. \label{ftrl-step}
\STATE Efficiently sample a $k$-set $S_t$ with inclusion probability vector $\bm{p}_t$ using Madow's sampling scheme (Algorithm \ref{uneq}).
\STATE Receive the reward function $f_t$ and set  $\bm{g}_t \gets \core{\alpha}{f_t}.$
\ENDFOR
\end{algorithmic}
\end{algorithm}

\begin{theorem}[Augmented regret bound for \textsc{SCore}] \label{main-thm}
Let each of the reward functions $\{f_t\}_{t\geq 1}$ be $\alpha$-admissible, non-negative, and bounded above by $M.$
The \onscore policy, described in Algorithm \ref{monotone-reward-algo}, achieves the following learning guarantee: 
		\begin{eqnarray} \label{regret-guarantee1}
		\frac{k}{N\alpha}\sum_{t=1}^Tf_t([N]) - \sum_{t=1}^T \mathbb{E}[f_t(S_t)] \leq 4M\sqrt{kT\ln\frac{N}{k}}.
	\end{eqnarray}
	The metric on the left-hand side of the above bound is called  $\alpha$-augmented regret.
\end{theorem}
\begin{proof}
From the static regret upper bound given by Theorem \ref{lin-reward}, we have:
	\begin{eqnarray} \label{sum-reward}
		\max_{\bm{p}^* \in \Delta^N_k} \sum_{t=1}^T \langle \bm{g}_t, \bm{p}^*\rangle - \sum_{t=1}^T \langle \bm{g}_t, \bm{p}_t \rangle \leq  2G\sqrt{2kT\ln\frac{N}{k}},
	\end{eqnarray}
	where $\bm{p}_t$ denotes the vector of inclusion probabilities on round $t,$ and $G$ is a uniform upper bound to the $\ell_2$-norms of the vectors $\{\bm{g}_t\}_{t=1}^T.$
Since each of the reward functions $\{f_t\}_{t=1}^T$ is assumed to be $\alpha$-admissible, and the vector $\bm{g}_t$ is chosen to be an $\alpha$-admissible vector, using Proposition \ref{core-bd}, we can set $G :=\alpha M \sqrt{2}.$ Furthermore, from the defining properties of the $\alpha$-\textsc{Core}, given in Eqns.\ \eqref{core-def1} and \eqref{core-def2}, we have:
	\begin{eqnarray} \label{est1}
		\mathbb{E}[f_t(S_t)] \geq \frac{1}{\alpha} \mathbb{E}[\sum_{i\in S} \bm{g}_{t,i}\mathds{1}(i \in S_t)] = \frac{1}{\alpha}\langle \bm{g}_t, \bm{p}_t \rangle, ~~\forall t \geq 1,
	\end{eqnarray}
	and
	\begin{eqnarray*}
		f_t([N]) = \sum_{i \in [N]} \bm{g}_{t,i}, ~~\forall t \geq 1.
	\end{eqnarray*}
	Hence,
	\begin{eqnarray} \label{est2}
		\sum_{t=1}^T f_t[N] = \sum_{i \in [N]}\sum_{t=1}^T  \bm{g}_{t,i} &\stackrel{(a)}{\leq}& \frac{N}{k} \max_{\bm{p}^* \in \Delta^N_k} \sum_{t=1}^T\langle \bm{g}_t, \bm{p}^*\rangle,
	\end{eqnarray}
	where in the inequality (a), we have 
	lower bounded the maximum over $\Delta^N_k$ with the feasible vector $\bm{p}\equiv\frac{k}{N}\bm{1}_{N}.$
	Finally, upon combining the bounds \eqref{sum-reward}, \eqref{est1}, and \eqref{est2}, we obtain the following upper bound on the augmented regret achieved by the \onscore policy: 
	\begin{eqnarray*} \label{regret-guarantee}
		\frac{k}{N\alpha}\sum_{t=1}^Tf_t([N]) - \sum_{t=1}^T \mathbb{E}[f_t(S_t)] \leq 4M\sqrt{kT\ln\frac{N}{k}}.
	\end{eqnarray*}
	\end{proof}
	\textbf{Note:} The \onscore policy recovers the standard regret bound \eqref{reg-bd-lin} for linear rewards (where $\alpha=1$) by choosing the benchmark $\bm{p}= \mathds{1}(S^*)$ after Eqn.\ \eqref{est1}, where $S^*$ is the optimal offline static $k$-set. 
	\paragraph{Discussion:}
	Recall that, in the definition of the standard regret, the offline benchmark is taken to be the cumulative reward accrued by the optimal \emph{fixed} offline action in hindsight \citep{zinkevich2003online}. Consequently, this benchmark could be weak if the reward functions change drastically over time. In order to address this fundamental issue, different variants of the standard regret, such as adaptive regret \citep{mokhtari2016online} and dynamic regret \citep{hazan2009efficient}, have been proposed, which allow some temporal variation of the offline benchmark. On the other hand, in augmented regret \eqref{regret-guarantee1}, the offline benchmark is taken to be a fraction $\eta \equiv \nicefrac{k}{N\alpha}$ of the maximum cumulative reward obtained by an omnipotent policy that selects \emph{all} $N$ elements at \emph{all} times, thus bypassing the issue discussed above. Furthermore, the benchmark in augmented regret is easy to compute compared to the benchmark in standard regret, which is often \textbf{NP-hard} to evaluate. The inverse relationship between $\eta$ and $\alpha$ signifies the (information-theoretic or computational) hardness for learning the rewards. To obtain a tighter bound, we want the parameter $\alpha$ to be as close to unity as possible. However, a given sequence of reward functions might not be $\alpha$-admissible for smaller values of $\alpha.$ Furthermore, even if the reward functions are known to be $\alpha$-admissible, finding an $\alpha$-admissible vector could be computationally intractable. The following section discusses this issue.

\subsection{Example of Admissible Rewards and Efficient Computation of Admissible Vectors} \label{admissible-rewards}
We now briefly outline our results on the characterization of the class of admissible rewards. Due to space constraints, we have moved the detailed discussions and derivations to Appendix \ref{reward-example}. We provide a number of examples of admissible reward functions which can be efficiently learned using the \onscore policy. Some highlights of our findings are listed below:

\begin{enumerate}
	\item An important class of $1$-admissible functions is the class of all submodular functions (see Appendix \ref{submod}). As an illustration, we consider the \textsc{Budgeted Assignment} Problem, where an optimal subset of $k$ webpages \emph{and} an optimal assignment of ads to the selected webpages need to be decided simultaneously in an online fashion (see Example \ref{budget-assignment} in Appendix \ref{reward-example}).
\item In many settings, the reward functions often have a natural description in terms of the solution of a combinatorial optimization problem. Hence, for maximizing these reward functions using the \onscore policy, one needs an efficient algorithm to compute admissible vectors. 
As an illustration, Section \ref{whitebox} shows how admissible vectors can be efficiently computed for the \textsc{Minimum-Cost Bipartite Matching} problem by utilizing dual LPs.  
\item In Sections \ref{non-submod1} and \ref{approx}, we consider the problem of efficiently maximizing a class of non-submodular reward functions. Theorem \ref{avg-submod} gives a sufficient condition for the \emph{Shapley value} of any reward function, possibly non-submodular, to be included in the core. In Definition \ref{dict-def}, we introduce a new notion, called the \emph{$m$-dictatorship}. Proposition \ref{dictator-prop} shows that a non-negative $m$-dictatorial set function, bounded above by $M$, is $\nicefrac{M}{m}$-admissible. In Example \ref{css}, we apply this result to the online \textsc{Sparse Regression} problem. Finally, Proposition \ref{g-construct} gives an efficient construction of an $\nicefrac{M}{m}$-admissible vector for any $m$-dictatorial reward function. 
\item Section \ref{rho-submod} considers a natural extension of submodularity, called $\rho$-submodularity. Using a greedy construction, Theorem \ref{approx-submod-core} shows that any $\rho$-submodular function is $\nicefrac{1}{\rho}$-admissible. Example \ref{batchbayes} applies this result to a Bayesian optimization problem. 
\end{enumerate}

\subsection{Comparison between the Augmented and Standard Regret for Submodular functions} As illustrated above, the augmented regret bound \eqref{regret-guarantee1} holds for a broad class of reward functions, including the class of all submodular functions, for which we have $\alpha=1$. As we explain below, in certain regimes, the augmented regret bound \eqref{regret-guarantee1} also implies stronger standard regret bounds for submodular functions compared to the best-known regret bounds in the literature.
 
 Since the reward functions are monotone non-decreasing, for any $k$-set $S^*$ (including the best offline $k$-set), we have $f([N]) \geq f(S^*).$ Since $\alpha=1$ for submodular functions (Theorem~\ref{submod-core}), the augmented regret bound in Theorem~\ref{main-thm} immediately yields a $(k/N)$-approximate $O(\sqrt{T})$ standard regret bound, which is achieved by the \onscore policy in near-linear time. In particular, if $k/N \geq 1-e^{-1}\approx 0.63$, this bound improves upon the standard $(1-e^{-1})$-approximate static regret bound for monotone submodular functions~\citep[Theorem 8, 9]{streeter2007onlinereport}. To the best of our knowledge, \onscore is the first efficient online policy that improves upon the standard $(1-e^{-1})$-approximate standard regret bound for submodular functions, even in this restricted regime. This result, in turn, also settles one of the open problems raised by \citet[Section 5]{streeter2007online} by showing that proving sublinear $(1-e^{-1})$-approximate static regret lower bound for submodular functions achievable by efficient policies for all $k\in[N]$ is infeasible.

The bound $f([N]) \geq f(S^*)$, used in the previous paragraph, might be loose if $k$ is small. In this case, if we consider submodular set functions with a positive curvature, the static regret bound implied by the \onscore policy gets even better. Recall from \cite{bian2017guarantees} that the generalized curvature of a non-negative monotone set function $f$ is defined to be the smallest scalar $c$ such that for all $A \subseteq B \subseteq [N]\setminus j$, we have 
 $ f(B \cup j) - f(B) \geq (1-c)(f(A \cup j)- f(A)). $
If the function $f$ is submodular, this immediately implies the following bound for any subset $S$.
\begin{align} \label{fun-bd}
    (1-c)\sum_{j \in S} f(j) \leq f(S) \leq \sum_{j \in S} f(j).
\end{align}
Using the estimate \eqref{fun-bd} in the static regret bound in Theorem~\ref{lin-reward}, we immediately conclude that the \onscore policy yields a $(1-c)$-approximate $O(\sqrt{T})$ static regret bound by taking $g_{t,j} = f_t(j), \forall j \in [N]$. This bound can be generalized for $\gamma$-weakly submodular functions~\citep{https://doi.org/10.48550/arxiv.2004.14650} as well, where one can similarly derive a $\gamma(1-c)$-approximate static regret guarantee with the \onscore policy.
	The following result gives a basic lower bound that shows that the worst-case augmented regret of any online policy is non-negative.
	\begin{theorem}[Lower bound on Augmented Regret] \label{monotone-reward-lb}
For any (randomized) online subset selection policy $\pi$,
 there exists a sequence of monotone non-negative linear (and hence, $1$-admissible) reward functions $\{f_t\}_{t \geq 1}$ such that its $1$-augmented regret   
 is non-negative, i.e.,
\begin{eqnarray} \label{perf_metric}
	    \frac{k}{N}\sum_{t=1}^T f_t([N]) - \sum_{t=1}^T \mathbb{E}_{\pi}f_t(S_t) \geq 0
\end{eqnarray}
	\end{theorem}
	The proof uses a probabilistic argument. See Appendix \ref{monotone-reward-lb-proof}  for the proof of Theorem \ref{monotone-reward-lb}.
	

\section{Optimistic Subset Selection for Submodular functions with Hints} \label{learningwithhints}
Recall that all submodular functions are known to be $1$-admissible (Theorem \ref{submod-core}), and hence, they admit an $O(\sqrt{T})$-augmented regret bound, thanks to Theorem \ref{main-thm}. We now extend the \onscore policy to a setting  
where the learner has access to some untrusted additional information or hint regarding the current reward function. Our objective is to judiciously use the hint provided to improve upon the augmented regret bound \eqref{regret-guarantee1}, if possible. In the context of online linear optimization, the problem of learning with hints has previously been considered by \citet{bhaskara2020online}, where the 
the quality of hint is measured in terms of the correlation between the hint and the true reward vectors. \citet{mhaisen2022optimistic} studied online linear optimization problems, where the hint quality is measured in terms of the $\ell_1$-norm of the difference between the hint and the true reward vectors. Since the \onscore policy reduces the subset selection problem to online linear optimization, our goal is to exploit this connection to extend the framework of learning with hints to the online subset selection problem. 
\paragraph{Problem Set up:}

Formally, we consider a setting where, on each round $t$, the learner is presented with an additional modular hint function $h_t:2^{[N]}\to\mathbb{R}$ for the (yet unrevealed) online submodular reward function $f_t$ 
\footnote{Recall that, a set function $h:2^{[N]} \to \mathbb{R}$ is called modular if it can be expressed as $h(S)=\sum_{i \in S}h_i, \forall S \subseteq [N]$ for some set of $N$ real coefficients $(h_i, 1\leq i \leq N).$}. The quality of the hint is measured in terms of their total variation distance, \emph{i.e.,} the $\sup$-norm of the difference between the hint and the reward functions:
\begin{eqnarray}
    \label{functionDistance}
   \textsf{Distance}(h_t, f_t)\equiv \max_{S\subseteq[N]}|f_t(S) - h_t(S)|.
\end{eqnarray}
We make \emph{no} assumptions about how the hints are generated or their quality. For example, the hints could be generated by a black box ML module that does not come equipped with any formal \emph{a priori} quality guarantee. We now show that, in this setting, the \onscore policy can be suitably extended to attain a dimension-free optimistic augmented regret bound of the form:
\begin{eqnarray}
    \label{functionDistanceRegret}
    O\left(\sqrt{\sum_{t = 1}^T\textsf{Distance}^2(h_t, f_t)}
\right). 
\end{eqnarray}
Clearly, the optimistic augmented regret bound \eqref{functionDistanceRegret} would substantially improve upon the worst-case $O(\sqrt{T})$ bound given in Theorem \ref{main-thm} if the hint functions closely approximate the reward functions. Otherwise, we still retain the $O(\sqrt{T})$ augmented regret guarantee of Theorem \ref{main-thm} (assuming the reward and the hint functions to be bounded). In order to establish the augmented regret bound \eqref{functionDistanceRegret}, we need the following technical lemma that lower bounds the distance between a submodular and a modular function in terms of the $\ell_1$ distance between any two vectors in their respective cores.   
\begin{lemma}
    \label{supermodularHintBound}
    Let $f:2^{[N]}\to\mathbb{R}$ be a submodular function and $h:2^{[N]}\to\mathbb{R}$ be a modular function. Let $\vec{f}\in\mathcal{B}_f$ be any vector belonging to the $1$-core (\emph{i.e.,} the Base polytope) of $f,$ and $\vec{h}$ be the coefficient vector of the modular hint function $h$. Then we have that
    \begin{align}
        ||\vec{f} - \vec{h}||_1\le 3\max_{S\subseteq[N]}|f(S) - h(S)|\equiv3 \textsf{Distance}(f,h) .
    \end{align}
\end{lemma}
See Appendix \ref{supermodularHintBoundproof} for the proof of Lemma \ref{supermodularHintBound}. 
We now use the \onscore policy by replacing the \textsc{FTRL} policy with an optimistic version of the FTRL policy (\textsc{OFTRL}) in the online linear optimization step \ref{ftrl-step}. The \textsc{OFTRL} policy uses a sequence of strongly convex adaptive regularizers $\{r_t(\cdot)\}_{t \geq 1}$ to obtain a tight optimistic regret bound. The pseudocode of the optimistic subset selection policy is given below in Algorithm \ref{oftrlHints}. 
Notice that the policy is computationally efficient as the vectors $\vec{f}_t$ and $\vec{h}_t$ can be efficiently computed in linear time. 
\begin{algorithm}
    \caption{\onscore with the Optimistic Follow The Regularized Leader (OFTRL) policy}\label{oftrlHints}
    \begin{algorithmic}[1]
        \STATE \textbf{Input:} Sequence of submodular rewards $\{f_t\}_{t \geq 1}$ and modular hint functions $\{\bm{h}_t\}_{t \geq 1}$
        \STATE \textbf{Output:} Sampled $k$-sets $\{S_t\}_{t\geq 1}$
        \STATE $\vec{\Theta}_0\gets \vec{0}, \sigma = 1/k$.
        \FOR{every round $t$}
            \STATE $\vec{p}_t\gets \underset{\vec{p}\in\Delta^N_k}{\mathrm{argmax}}\left[-\sum_{\tau=1}^{t-1}r_{\tau}(\vec{p}) + \left\langle \vec{p}, \vec{\Theta}_{t - 1} + \vec{h}_t\right\rangle\right]$.
             \textcolor{blue}{// See Section \ref{p-comp} in the Appendix for a  Frank-Wolfe-based implementation with a round wise complexity of $O(N \log N \log T)$}
          \STATE Efficiently sample a $k$-set $S_t$ according to the inclusion probability vector $\bm{p}_t$ using Madow's sampling scheme (Algorithm \ref{uneq}).
            \STATE Obtain the reward function $f_t$, and choose any vector $\vec{f}_t$ from the base polytope $\mathcal{B}_{f_t}$.
            \STATE Update $\vec{\Theta}_{t}\gets \vec{\Theta}_{t - 1} + \vec{f}_t$.
            \STATE $\delta_t\gets ||\vec{f}_t - \vec{h}_t||_2^2$. \label{delta-def}
            \STATE Define the strong convexity parameter $\sigma_t := \sigma(\sqrt{\sum_{\tau=1}^t\delta_{\tau}} - \sqrt{\sum_{\tau=1}^{t-1}\delta_{\tau}})$. \label{sigma-def}
            \STATE Define $\sigma_t$-strongly convex regularizer $r_t(\vec{x}) := \frac{\sigma_t}{2}||\vec{x} - \vec{p}_t||_2^2$
        \ENDFOR
    \end{algorithmic}
\end{algorithm}
The following Theorem gives an optimistic augmented regret bound for the above policy.
\begin{theorem}
    \label{oftrlBound}
    Algorithm \ref{oftrlHints} yields the following augmented regret bound for the online subset selection problem for any sequence of submodular functions $\{f_t\}_{t\geq 1}$ with modular hints $\{h_t\}_{t \geq 1}$: 
    \begin{eqnarray}
        \dfrac{k}{N}\sum_{t = 1}^T f_t([N]) - \sum_{t = 1}^T f_t(S_t)\le 12k\sqrt{\sum_{t = 1}^T \textsf{Distance}^2(f_t, h_t).}
    \end{eqnarray}
\end{theorem}
See Appendix \ref{OptimisticBdProof} for the proof. Note that the above bound is \emph{dimension-free}, \emph{i.e.,} independent of $N.$ 

\textbf{Note:} It is easy to verify that Theorem \ref{oftrlBound} holds (with an identical proof) for the more general class of $1$-admissible reward functions.
\section{Related Works} \label{related-works}
\subsection{Online Set Function Maximization}
Most of the previous works in online set function maximization have been restricted to the case when the reward functions are submodular. Numerous works have studied online submodular maximization problems since the seminal paper by~\cite{streeter2007online}, which greedily runs $k$ expert policies at every round to sample a set of $k$ elements. 
\cite{roughgarden2018optimal} studied the non-monotone submodular optimization problem and provided an online double greedy strategy that achieves a $\nicefrac{1}{2}$-approximate regret bound. Recently, \cite{harvey2020improved} improved the static regret bounds for both problems. Furthermore, both problems have been explored with the notion of tracking regret by~\cite{matsuoka2021tracking}. 
The study of online non-submodular set function maximization has been driven by the online sparse linear regression problem~\citep{foster2016online, kale2017adaptive}. \cite{kale2017adaptive} provided an efficient algorithm for the problem in the adversarial setting by leveraging the greedy strategy proposed by \cite{streeter2007online} and incorporating weak submodularity~\citep{boutsidis2015greedy} in their utility set functions. The subset-selection problem for linear reward functions arises in online caching, which has been recently studied by \citet{sigmetrics20, georgios-no-regret, paria2021}. 
%


\subsection{Weak Submodularity}
In many real-world problems ensuring utility functions are submodular can be pretty restrictive. For example, the problems of dictionary selection~\citep{cevher2011greedy} and column subset selection~\citep{boutsidis2009improved,altschuler2016greedy} are known to have non-submodular utility functions. To address this issue, several works have considered different notions of weak submodularity. The idea of submodularity ratio and, consequently, weak submodularity was introduced by~\cite{10.5555/3104482.3104615} in the context of the subset selection problem. The concept of weak submodularity has also been studied by relaxing the notion of decreasing marginal gains by a factor of $\gamma\in[0,1]$ under names such as \textit{inverse curvature}~\citep{pmlr-v84-bogunovic18a}, \textit{DR ratio}~\citep{pmlr-v80-kuhnle18a} and \textit{generic submodularity ratio}~\citep{nong2019maximize}. \cite{https://doi.org/10.48550/arxiv.2004.14650} extended the notion of weak submodularity for non-monotone functions and introduced the notion of \textit{local submodularity ratio}.  \citet{halabi2018combinatorial} introduced the notion of \emph{$\rho$-submodularity}, and~\cite{halabi2020optimal} introduced the notion of \emph{weak diminishing return (DR) submodularity}, both of which are closely related to weak submodularity and have been frequently employed in the recent literature on non-submodular optimization. Recently, \cite{pmlr-v178-thiery22a} further refined the concept of the submodularity ratio defined by \cite{10.5555/3104482.3104615} for monotone functions.
\subsection{Approximate Core in Cooperative Game Theory} \cite{shapley1971cores} proved that convex games~(\emph{i.e.,} games with a submodular cost function) have non-empty cores. However, the core of a cooperative game could be empty for many interesting classes of reward functions \citep{maschler2020game}. Various approximations of the core have been introduced to overcome this problem. 
Examples of additive approximations to the core of cooperative games are \textit{least core}~\citep{maschler1979geometric} or \textit{nucleolus}~\citep{schmeidler1969nucleolus}. Recently, \cite{vazirani2022general} provided an $\alpha$-multiplicative approximation of the core in the context of a general graph-matching game for $\alpha=2/3$. Similarly, \cite{munagala2022approximate} proposed a multiplicative approximation of the core for the committee selection problem.

\subsection{Learning with Hints}
\citet{hazan2007priorknowledge} considered an example of the online linear optimization problem where the learner knows the first coordinate $\vec{c}_{t,1}$ of the cost vector $\vec{c}_t\in[-1, 1]^n$ for all $t$. They showed that if the set of feasible actions is the Euclidean ball, then a regret bound of $O(\alpha^{-1} \log T)$ is achievable. \citet{dekel2017hint} achieved a similar regret bound when the decision space is \textit{uniformly convex} and symmetric around the origin, and there are unit-length hint vectors $\vec{h}_t$ which satisfy $\langle\vec{h_t}, \vec{c_t} \rangle \ge\alpha||\vec{c}_t||_2$. \citet{bhaskara2020online} came up with an algorithm that adapts to the quality of the correlation of the hint vector $\vec{h}_t$ and the cost vector $\vec{c}_t$; they achieved a regret bound of the form $O\left((1 + B)/\alpha\cdot\log(1 + T - B)\right)$, where $B$ is the number of time steps when the hints are of \textit{bad} quality, i.e $\langle \vec{c}_t, \vec{h}_t\rangle < \alpha||\vec{c}_t||_2^2$. 
\citet{rakhlin2013optimization} 
designed an online algorithm such that if the learner receives hint vector $\vec{M}_t$ in round $t$, then the regret of the algorithm is of the order $O\left(\sum_{t = 1}^T||\vec{\nabla}_t - \vec{M}_t||_*^2\right),$ where $\vec{\nabla}_t$ is the gradient of the loss function on round $t$. 

\section{Concluding Remarks and Open Problems} \label{conclusion}
In this paper, we proposed an efficient online subset selection policy and its optimistic version, called \textsc{SCore}, for maximizing a broad class of reward functions. The \onscore policy uses the concept of approximate core from the cooperative game theory literature to obtain modular approximations to the reward functions, and guarantees an $O(\sqrt{T})$ $\alpha$-\emph{augmented-regret}. We now discuss some related future research directions.  
First, an interesting open problem is to extend the \onscore policy to the bandit feedback setting, where only the reward value for the selected subset is revealed to the learner at the end of each round. \edit{Second, it would be interesting to generalize our results to matroids other than the uniform matroid setting that we study in this paper \citep{golovin2014online}.}
\edit{Third, note that the \onscore policy and its optimistic version work with \emph{any} appropriate admissible vectors. It is an interesting open problem to see if the stronger regret bounds can be derived by optimizing the choice of the admissible vectors used in the \onscore policy.} 
Fourth, the lower bound in Theorem \ref{monotone-reward-lb} is pretty basic. It would also be interesting to establish a lower bound on the $\alpha$-regret that either matches the upper bound or exhibits a non-trivial dependence on the horizon length for a class of reward functions.
Fifth, it would be nice to characterize the trade-off between the computational complexity of computing an $\alpha$-admissible vector as a function of the approximation parameter $\alpha$ for a broad class of reward functions. 
\newpage
\bibliography{bibmobility}
\newpage
\appendix
\section{Madow's Sampling Algorithm} \label{madow-section}
The following pseudocode outlines Madow's algorithm \citep{madow1949theory}, that randomly samples a subset $S$ consisting of $k$ elements \emph{without replacement} from a universe $[N]$ containing $N$ elements such that element $i \in [N]$ is included in the sample $S$ with a pre-specified inclusion probability $p_i, i \in [N].$ The marginal inclusion probabilities are assumed to satisfy the necessary conditions \eqref{necsuf1} and \eqref{necsuf2}.
\begin{algorithm} 
\caption{Madow's Sampling Scheme}
\label{uneq}
\begin{algorithmic}[1]
\REQUIRE A universe $[N]$ of size $N$, the cardinality of the sampled set $k$, and a  marginal inclusion probability vector $\bm{p} = (p_1, p_2, \ldots, p_N)$ satisfying conditions \eqref{necsuf1} and \eqref{necsuf2}.
\ENSURE A random $k$-set $S$ with $|S|=k$ such that, $\mathbb{P}(i \in S)=p_{i}, \forall i\in [N]$ 

\STATE Define $\Pi_0=0$, and $\Pi_i= \Pi_{i-1}+p_{i}, \forall 1\leq i \leq N.$
\STATE Sample a uniformly distributed random variable $U$ from the interval $[0,1].$
\STATE $S \gets \emptyset$
\FOR {$i\gets 0$ to $k-1$}
\STATE Select the element $j$ if $\Pi_{j-1} \leq U + i < \Pi_j.$ 
\STATE $S \gets S \cup \{j\}.$
\ENDFOR
\RETURN $S$
\end{algorithmic}
\end{algorithm}
\paragraph{Correctness:} It can be easily verified that Madow's sampling scheme, described in Algorithm \ref{uneq}, outputs a random set $S, |S|=k,$ such that $\mathbb{P}(j \in S) = p_j, \forall i \in [N].$ To see this, by the feasibility condition of the inclusion probabilities, we have $\Pi_{N}= \sum_{j=1}^Np_j=  k.$ Hence, Algorithm \ref{uneq} samples exactly $k$ elements. Furthermore, since $\Pi_j - \Pi_{j-1} = p_j \leq 1,$ for any realization of the uniform r.v. $U,$ there exists a unique integer $i_j^*$ such that $ \Pi_{j-1} \leq U+i_j^* \leq \Pi_j.$  
Hence, $\forall j \in [N],$ we have that
\begin{eqnarray*}
	\mathbb{P}(j \in S) = \mathbb{P}(\Pi_{j-1}\leq U+i^*_j \leq \Pi_j) = \mathbb{P}(\Pi_{j-1}- i^*_j\leq U \leq \Pi_j- i^*_j) = \Pi_j-\Pi_{j-1} = p_j.
\end{eqnarray*}
This completes the proof of the correctness of Madow's sampling algorithm. $\blacksquare$


\section{Proof of Theorem \ref{ex-bs}} \label{ex-bs-proof}
\begin{proof}
Consider the following primal LP, which corresponds to $\core{\alpha}{f}.$ \\

$\mathcal{P}:$\\
\begin{eqnarray*}
	\textrm{maximize}~~~ 0,
\end{eqnarray*}
subject to
\begin{eqnarray} \label{primal}
	\sum_{i \in S} g_i &\leq& \alpha f(S),~ \forall S \subseteq [N], \label{pr1}\\
	\sum_{i \in [N]}g_i &=& f([N]). \label{pr2}
\end{eqnarray}
Associating the dual variable $\delta_S \geq 0$ to the constraint corresponding to the subset $S$ in \eqref{pr1} $\forall S \subseteq [N],$ and a dual variable $\mu \in \mathbb{R}$ to the constraint \eqref{pr2}, we can write down the dual LP $\mathcal{D}$ corresponding to the primal LP $\mathcal{P}$ as follows. \\

$\mathcal{D}:$ 

\begin{eqnarray} \label{dual}
	\textrm{minimize}~~~ \alpha \sum_{S \subseteq [N]} \delta_S f(S) + \mu f([N]), 
\end{eqnarray}
subject to 

\begin{eqnarray}
	\sum_{S: i \in S} \delta_S+ \mu &=&0, ~~ \forall i \in S, \label{d1}\\
	\delta_S &\geq& 0, ~~ \forall S \subseteq [N] \label{d2}. 
\end{eqnarray}
It can be seen that the dual $\mathcal{D}$ is feasible, and its optimal value is  upper-bounded by $0$ (take, \emph{e.g.,} the feasible solution $\delta_S =0, \forall S \subseteq [N]$, and $\mu = 0.$) Hence, from the Strong duality theorem, $\core{\alpha}{f}$ is non-empty if and only if the objective value of the dual $\mathcal{D}$ is non-negative for all dual feasible solutions. In other words,
\begin{eqnarray}
	&&\core{\alpha}{f} \neq \emptyset \nonumber \\
	 &\Longleftrightarrow& \alpha \sum_{S \subseteq [N]} \delta_S f(S) \geq - \mu f([N]) ~\forall \bm{\delta}, \mu~ \textrm{satisfying}~ \eqref{d1} \textrm{ and } \eqref{d2}. \\
	&\Longleftrightarrow& \alpha \sum_{S \subseteq [N]} \delta_S f(S) \geq - \mu f([N]) ~\forall \bm{\delta}, \mu~ \textrm{satisfying}~ \eqref{d1},\eqref{d2}~ \textrm{and} ~ \mu < 0,
\end{eqnarray}
where the last equivalence follows from the observation that since $f$ is non-negative, it is sufficient to check the inequality for strictly negative $\mu$ only. Finally, rescaling the variables $\delta_S \gets \frac{\delta_S}{-\mu}, \forall S \subseteq [N],$ we conclude that 
\begin{eqnarray*}
	\core{\alpha}{f} \neq \emptyset &\Longleftrightarrow& \alpha \sum_{S \subseteq [N]} \delta_S f(S) \geq f([N])
\end{eqnarray*}
for any set of variables $\bm{\delta}$ satisfying
\begin{eqnarray*}
	\sum_{S: i \in S} \delta_S &=& 1, ~\forall i \in S\\
	\delta_S &\geq& 0, ~\forall S \subseteq [N].
\end{eqnarray*}
This proves Theorem \ref{ex-bs}.
\end{proof}
\section{Proof of Proposition \ref{core-bd}} \label{core-bd-proof} 
\begin{proof}
Using the definition of $\alpha$-\textsc{Core} and the fact that $\bm{g} \in \core{\alpha}{f}$, we have:
\begin{eqnarray} \label{g-bd1}
 g_i \leq \alpha f(\{i\})\leq \alpha M, ~~\forall i \in [N].
\end{eqnarray} 
Similarly, we have 
\begin{eqnarray} \label{g-bd2}
	\sum_{j \in [N]\setminus \{i\}} g_j \leq \alpha f([N]\setminus \{i\})\leq \alpha M.
\end{eqnarray}
	Using \eqref{g-bd2} and the definition of $\alpha$-\textsc{Core} again, we can write 
	\begin{eqnarray} \label{g-bd3}
	\alpha M +g_i \geq 	g_i + \sum_{j \in [N]\setminus \{i\}} g_j = f([N]) \geq 0.
	\end{eqnarray}
	Combining \eqref{g-bd1} and \eqref{g-bd3}, we have the following upper bound for the absolute value of each component of the vector $\bm{g}:$
	\begin{eqnarray} \label{g-bd0}
	|g_i| \leq \alpha M, ~~\forall i \in [N].
	\end{eqnarray}
	 Now define the index sets $P$ and $N$ as $P=\{i: g_i \geq 0\}$ and $N=\{i: g_i<0\}.$ Using the definition of $\alpha$-\textsc{Core} again, we have:
	\begin{eqnarray} \label{g-bd4}
		\sum_{i\in P} g_i \leq \alpha M. 
	\end{eqnarray}
	Similarly,
	\begin{eqnarray*} 
		\alpha M + \sum_{i \in N}g_i \geq \sum_{i \in P} g_i + \sum_{j \in N}g_j = f([N]) \geq 0.
	\end{eqnarray*}
	This yields
	\begin{eqnarray} \label{g-bd5}
		\sum_{j \in N} g_j \geq -\alpha M. 
	\end{eqnarray}
	Finally,
	\begin{eqnarray*}
		G^2 \leq ||\bm{g}||_2^2 &=& \sum_{i \in P}g_i^2 + \sum_{j \in N}g_j^2 \\
		&\stackrel{(a)}{\leq} & \alpha M \sum_{i \in P } g_i - \alpha M \sum_{j \in N}g_j\\
		&\stackrel{(b)}{\leq} & 2\alpha^2 M^2,
	\end{eqnarray*}
	where (a) follows from \eqref{g-bd0}, and the definition of the index sets $P$ and $N$ and (b) follows from \eqref{g-bd4} and \eqref{g-bd5}.
\end{proof}

\begin{proposition} \label{beta-submod-norm}
Consider a monotone $\rho$-submodular reward function $f$ taking values in the interval $[0,M].$ Then, $\exists \bm{g} = (g_1, g_2,\dots, g_N)\in\core{\rho^{-1}}f$ such that $||\bm{g}||_2 \leq M.$ 
\end{proposition}

\begin{proof}
Take $\bm{g}$ to be the same vector as given in Theorem \ref{approx-submod-core}, \emph{i.e.},
\begin{align}
    g_i := f(\{1, 2, \dots, i\}) - f(\{1, 2, \dots, i-1\}), ~ \forall i \in [N].
\end{align}
Theorem \ref{approx-submod-core}, proved in the following section, shows that $\bm{g}\in\core{\alpha}f$. Furthermore, notice that $g_i\geq0,\forall i\in[N],$ as $f$ is monotone non-decreasing. Hence, we have
\begin{eqnarray*}
    \norm{\bm g}^2_2\leq \norm{\bm g}^2_1=(\sum_{i=1}^N g_i)^2 = f^2([N])\leq M^2.
\end{eqnarray*}
\end{proof}

\section{Some Examples of Admissible Reward Functions} \label{reward-example}
In this section, we give several examples of admissible reward functions that often arise in practice. Since the only non-trivial part of the \onscore policy is the computation of an $\alpha$-admissible vector for some $\alpha \geq 1,$ we exclusively focus on this problem in this section. 
\subsection{Submodular Rewards} \label{submod}
Informally speaking, submodular functions are set functions that have the \emph{diminishing return} property as formalized below. 
\begin{definition}[Submodular function]
	A function $f: 2^{[N]}\to \mathbb{R}$ is called submodular if $\forall B \subseteq [N], A \subseteq B, i \in B^c:$
\begin{eqnarray} \label{diminishing}
f(B\cup \{i\}) - f(B) \leq f(A\cup \{i\})- f(A). 	
\end{eqnarray}
\end{definition}
\textbf{Examples:} Many objective functions arising in the field of machine learning and artificial intelligence are known to be submodular. See \citet{bilmes2022submodularity} for a recent survey. In the following, we give only some representative examples. 
\begin{enumerate}
	\item Let $\bm{X}= (X_1, X_2, \ldots, X_N)$ be a random vector. Then the Entropy function $H(\bm{X}_S)$ and the mutual information function $I(\bm{X}_S; \bm{X}_{S^c})$ are known to be monotone submodular.
	\item Consider a social network represented by a graph $G(V,E),$ all of whose nodes are initially in the inactive state. For a subset $S\subseteq V$, define the \emph{influence function} $\sigma(S)$ to be the expected number of active nodes at the end of an influence spread process where the subset $S$ was activated at the beginning. In a pioneering paper, \citet[Theorem 2.2, Theorem 2.5]{kempe2003maximizing} showed that under two standard influence spread processes (Independent Cascade and Linear Threshold), the influence function $\sigma(\cdot)$ is monotone submodular.
	
	\item Given a family of sets $\{U_v\}_{v \in V},$ the coverage function defined as $\textsc{Cover}(S):=|\cup_{i\in S} U_i|, S\subseteq V$ is monotone submodular. 	
\end{enumerate} 
The following result shows that the \onscore policy can efficiently learn the class of submodular functions. 
\begin{theorem}{\citep{shapley1971cores}} \label{submod-core}
	Submodular functions are $1$-admissible. Furthermore, for any permutation $\pi :[N]\to [N],$  define the marginal utility vector as $\Delta^\pi_{\pi(i)} \equiv f({\pi(1),\pi(2),\ldots, \pi(i)}) - f(\pi(1), \pi(2),\ldots, \pi(i-1)), \forall i\in [N].$ Then $\bm{\Delta}^\pi \in \core{1}{f}.$
\end{theorem}
Theorem \ref{submod-core} is a well-known result in cooperative game theory. We state and prove a generalized version of this result in Theorem \ref{approx-submod-core}. The above results imply that many practical online subset selection problems with cardinality constraints, including the online version of the sensor placement problem \citep{guestrin2005near} and the influence maximization problem in social networks with targeted interventions \citep{kempe2003maximizing} can be efficiently solved using \textsc{SCore}. The following examples give more nuanced applications of the \onscore policy. 
\begin{example}[Budgeted Assignment \citep{golovin2009online}] \label{budget-assignment}
	Assume that there are $N$ webpages, denoted by the set $[N],$ where one may buy ad spaces for displaying ads. However, due to budget constraints, we can afford to buy ad spaces in only a subset $S \subseteq [N]$ of $k$ webpages. If page $i$ is selected for advertising, we may choose to display any one ad from the pool of ads $P_i, i \in [N].$ Let $\Pi = (p_1, p_2, \ldots, p_N)$ be an assignment which assigns an ad $p_i \in P_i$ to each page $i$. The reward (\emph{e.g.,} the cumulative click-through rate) for the choice of webpages $S$ and the ad assignment $\Pi$ is given by the function $f(S, \Pi)$. For a fixed assignment $\Pi$, the reward function $f(\cdot, \Pi)$ is typically assumed to be monotone and submodular, and thus $1$-admissible by Theorem \ref{submod-core}. Hence, Theorem \ref{main-thm} implies that a sequence of reward functions $\{f_t(\cdot, \Pi)\}_{t \geq 1}$ can be efficiently learned by the \onscore policy. In ad exchanges, one is typically interested to learn \underline{both} the optimal subset of pages $S^* (|S^*|=k)$ to put ads on \underline{and} the optimal assignment of ads $\Pi^*$ to pages. Similar problems arise in news recommendations and sponsored searches. In the following, 
	we show how this more general problem can be efficiently solved by the \onscore policy. 
\end{example}

\subsubsection{Analysis of the \textsc{Budgeted Assignment Problem}}
The paper by \citet{golovin2009online} considers a particular case of the online assignment problem, discussed in Example \ref{budget-assignment}, with \emph{no} budget constraints. In this problem, one may place ads on \emph{all} of the $N$ web pages (\emph{i.e.,} $k=N$). Let $\mathcal{A}$ denote the set of all assignments of ads to webpages. Given a sequence of reward functions $\{f_t([N], \cdot)\}_{t \geq 1},$ their proposed online assignment policy, called \textsc{TGbandit}, selects a series of assignments $\{G_t\}_{t \geq 1},$ achieving the following $(1-\nicefrac{1}{e})$ regret guarantee:
\begin{eqnarray} \label{assignment-regret}
	\mathbb{E}\big[\sum_{t=1}^T f_t([N], G_t)\big] \geq (1-\frac{1}{e}) \max_{G \in \mathcal{A}} \big[\sum_{t=1}^T f_t([N], G) \big] - \tilde{\Theta}(N^{3/2}T^{1/4}\sqrt{\textsf{OPT}}),
\end{eqnarray}
where $\textsf{OPT}\equiv \max_{G \in \mathcal{A}}\sum_{t=1}^T f_t([N], G)$ is the cumulative reward accrued by the optimal static assignment. We can now combine the \onscore and \textsc{TGbandit} policy to solve the \textsc{Budgeted Assignment Problem} for an arbitrary budget constraint, where only $k$ of the $N$ webpages may be selected for advertisements. Our proposed policy is described in Algorithm \ref{budgeted-assignment-alg} below.
\begin{algorithm} 
\caption{Online policy for the \textsc{Budgeted Assignment Problem}}
\label{budgeted-assignment-alg}
\begin{algorithmic}[1]
\FOR {every round $t$:}
\STATE $G_t \gets \textsc{TGBandit}$  ~~\textcolor{blue}{// Selects an assignment on all $N$ webpages}
\STATE $S_t \gets \onscore$ ~~~~~~~~\textcolor{blue}{// Selects a subset of $k$ pages}
\STATE Feed the reward function $f_t([N], \cdot)$ to \textsc{TGBandit}
\STATE Feed the reward function $f_t(\cdot, G_t)$ to \onscore
\ENDFOR
\end{algorithmic}
\end{algorithm}

Algorithm \ref{budgeted-assignment-alg} runs an independent instance of the \textsc{TGbandit} policy, which determines a sequence of ad assignments for \emph{all} $N$ web pages on all rounds. Hence, the run of the \textsc{TGbandit} policy is \emph{not} affected by the concurrently running \onscore policy. On the other hand, the reward inputs to the \onscore policy are modulated with the choice of assignment made by the \textsc{TGbandit} policy.
\paragraph{Analysis:} From the augmented regret guarantee \eqref{regret-guarantee1} achieved by the \onscore policy, we have 
\begin{eqnarray}\label{reg-guarantee2}
	\mathbb{E}\big[\sum_{t=1}^T f_t(S_t, G_t) \big] \geq \frac{k}{N} \sum_{t=1}^T f_t([N], G_t) -\tilde{\Theta} (\sqrt{kT}).
\end{eqnarray}
Combining \eqref{assignment-regret} and \eqref{reg-guarantee2}, we have the following learning guarantee for Algorithm \ref{budgeted-assignment-alg}:
\begin{eqnarray}\label{budgeted-assignment-guarantee}
	\mathbb{E}\big[\sum_{t=1}^T f_t(S_t, G_t) \big] \geq (1-\frac{1}{e})\frac{k}{N}\textsf{OPT} - \tilde{\Theta}(kN^{1/2}T^{1/4}\sqrt{\textsf{OPT}}+ \sqrt{kT}),
\end{eqnarray}
where we recall $\textsf{OPT}\equiv \max_{G \in \mathcal{A}} \big[\sum_{t=1}^T f_t([N], G) \big]$ is the maximum cumulative reward accrued by the best static offline assignment when ads are displayed on \emph{all} $N$ web pages. It is worth pointing out that Algorithm \ref{budgeted-assignment-alg} and its analysis critically exploit that \onscore policy has a sublinear \emph{augmented regret}. The webpage selection algorithm and its analysis would have been much more complex had we used a subset selection policy with a static regret guarantee only (\emph{e.g., \citet{streeter2007online}}).

%
\subsection{White-box access to combinatorial rewards} \label{whitebox}
Given black-box oracle access to a submodular function, Theorem \ref{submod-core} gives an efficient method to compute a $1$-admissible vector. However, in many problems, the reward function is naturally defined as a solution to a non-trivial combinatorial optimization problem. Hence, invoking the evaluation oracle could be prohibitively expensive for these problems. Instead, the learner is provided access to the underlying combinatorial object itself (\emph{e.g.,} the graph). The following example, adapted from the classic paper by \citet{shapley1971assignment}, shows that a $1$-admissible vector for the Minimum-Cost Bipartite Matching problem can be computed efficiently. 

\begin{example}[Minimum-Cost Bipartite Matching] \label{assignment}
Consider a non-negative weighted complete bipartite graph $G(U,V, E,w)$, where the cost of the edge $(u,v) \in E$ is $w_{uv}.$ Let $k \leq 2 |U|$ be an even integer. For any subset of vertices $S_U \subseteq U, S_V \subseteq V$ consisting of $k/2$ elements each, let the value of the reward function $f(S_U \cup S_V)$ be given by the weight of the minimum cost perfect matching on the subgraph induced by the vertices $S \equiv S_U \sqcup S_V$. Using ideas from bipartite matching and duality theory, in the following subsection, we show that the above reward function $f$ is $1$-admissible. Furthermore, a $1$-admissible vector can be efficiently computed from the optimal solution to the dual of the Minimum Cost Matching LP.
\end{example}

 \subsubsection{Core of the Minimum-Cost Bipartite Matching Problem}\label{assignment-core}
  
 Consider the LP relaxation of the minimum cost perfect matching problem on the weighted complete bipartite graph $G(U,V, E, w),$ where the variable $x_{ij}, (i,j)\in E$ denotes the extent to which vertex $i$ is matched with vertex $j.$ It is well-known that the relaxed LP is Totally Unimodular and has an integral optimal solution \citep{schrijver1998theory}. The value of the minimum cost matching $f([N])$ for the entire bipartite graph is given by the optimal value of the following LP.
   \begin{eqnarray} \label{assignment_LP}
   	\min \sum_{(i,j)\in E} w_{ij}x_{ij}
   \end{eqnarray}
   s.t. 
   \begin{eqnarray}
   	\sum_{j: (i,j)\in E} x_{ij} &\geq& 1, ~~\forall i \in U \label{lp1}\\
   	\sum_{i: (i,j) \in E} x_{ij} &\geq& 1, ~~ \forall j \in V \label{lp2}\\
   	x_{ij}&\geq& 0,~~ \forall (i,j) \in E.
   \end{eqnarray}
    Associating a dual variable $u_i$ to the constraint \eqref{lp1} and a dual variable $v_j$ to the constraint \eqref{lp2} for all $i \in U, j \in V,$ we can write down the dual of LP \eqref{assignment_LP} as follows:
    \begin{eqnarray*}
    	\max \sum_{i \in U} u_i + \sum_{j \in V} v_j
    \end{eqnarray*}
    s.t.
    \begin{eqnarray}
    	u_i + v_j &\leq& w_{ij}, ~~\forall (i,j) \in E \label{dlp1} \\
    	u_i &\geq& 0, ~~ \forall i \in U \label{dlp2}\\
    	v_j &\geq & 0, ~~\forall j \in V \label{dlp3}.
    \end{eqnarray}
    Let $(\bm{u}^*, \bm{v}^*)$ be an optimal solution to the dual LP, which can be efficiently obtained using the Hungarian algorithm \citep{kuhn1955hungarian}. Define the components of a vector $\bm{g}$ as follows: 
    \begin{eqnarray*}
    	  g_k =\begin{cases} u_k^*, ~\textrm{if}~ k\in U \\
    	  v_k^*, ~ \textrm{if}~ k \in V. 
    	  \end{cases}
    \end{eqnarray*}
    We now show that the vector $\bm{g} \in \core{1}{f}.$ Since both the primal and dual LPs are feasible, strong duality implies that    
    \begin{eqnarray} \label{core-eq1}
    	f([N]) = \sum_{i \in U} u_i^* + \sum_{j \in V} v_j^* = \sum_{i \in U \sqcup V} g_i.
    \end{eqnarray}
       On the other hand, consider any perfect matching $\mathcal{M}$ on the nodes $S_U \sqcup S_V.$ From the dual feasibility of $(\bm{u}^*, \bm{v}^*)$ and using the dual constraint \eqref{dlp1}, we have 
    \begin{eqnarray*}
    	\sum_{(i,j) \in \mathcal{M}} w_{ij} \geq \sum_{i \in S_U} u_i^* 
+ \sum_{j \in S_V} v_j^*.
    \end{eqnarray*}
    Taking minimum over all perfect matchings $\mathcal{M}$ on $S\equiv S_U \sqcup S_V$, we have 
    \begin{eqnarray} \label{core-eq2}
    	\sum_{i \in S} g_i \leq f(S).
    \end{eqnarray}
    Equations \eqref{core-eq1} and \eqref{core-eq2}, taken together, shows that $\bm{g} \in \core{1}{f}.$ Interestingly, from the dual feasibility \eqref{dlp2} and \eqref{dlp3}, it also follows that the vector $\bm{g}$ is component-wise non-negative. Furthermore, from Proposition \ref{core-bd}, it follows that $||\bm{g}||_2 \leq Nw_{\max}/\sqrt{2}.$
  \paragraph{Note:} The above reward function arises in a two-player zero-sum game where the first player chooses two subsets $S_U \subseteq U$ and $S_V \subseteq V$ s.t. $|S_U|=|S_V|=k/2.$ Upon observing  the first player's choice, the second player chooses a perfect matching on the selected vertices and incurs a cost equal to the price of the matching. Hence, for any choice of the subsets, the reward obtained by the first player is equal to the cost of the minimum cost perfect matching on the selected vertices.

\subsection{Non-Submodular Rewards and Shapley Value} \label{non-submod1}
If the reward function is not submodular, the marginal utility vector $\bm{\Delta}^\pi$ for a given permutation $\pi,$ defined in Theorem \ref{submod-core}, might not lie in the core even if the core is non-empty. Fortunately, for some reward functions satisfying certain conditions, the \emph{Shapley value} of the reward function, defined below, is guaranteed to be included in the core.    
\begin{definition}[Shapley value]
	The Shapley value $\textrm{Sh}(f) $ is defined as $\textrm{Sh}_i(f) = \mathbb{E}\Delta^\pi_{\pi(i)}, \forall i \in [N]$ where the expectation is over all permutations on $[N]$ taken uniformly at random. 
\end{definition}
 Shapley value exists for any reward function. Although the problem of computing the Shapley value for arbitrary reward functions is \textbf{NP-Hard}, efficient quasi-Monte Carlo algorithms for estimating the Shapley value exist \citep{mitchell2022sampling}. Obviously, showing that the Shapley value is in the core is sufficient to prove the non-emptiness of the core (and hence, the $1$-admissibility of the reward function). Towards this, we recall the following result.

\begin{theorem}{\citep{izawa1998coalitional}} \label{avg-submod}
	The Shapley value of a function $f$ is included in the core if and only if for all $T \subseteq [N],$ the following holds 
	\begin{eqnarray*}
		\sum_{S \subseteq [N]}\sum_{i \in S\cap T}\frac{(|S|-1)!(N-|S|)!}{N!}[f_i(S)-f_i(S\cap T)] \leq 0,
	\end{eqnarray*}
	where $f_i(S)  \equiv f(S) - f(S \setminus \{i\})$ is the marginal contribution of the $i$\textsuperscript{th} element to the set $S \ni i.$ 
\end{theorem}

As an application of the above result, recall that a reward function $f$ is called \emph{average submodular} if the diminishing return property \eqref{diminishing} holds on the average, \emph{i.e.,} $\forall A, B \neq \emptyset, A\cap B =\emptyset:$
\[ f(A\cup B)- f(B) \leq \frac{1}{|B|}\sum_{i \in B} \big[f\big(A\cup (B\setminus \{i\})\big) - f\big(B\setminus \{i\}\big)\big]. \]
Theorem \ref{avg-submod} implies that average submodular functions have a non-empty core, and the Shapley value is in the core \citep{sprumont1990population}. Note that average submodular functions need not be submodular \citep{inarra1993shapley}.
\subsection{Non-Submodular Rewards and $\alpha$-admissibility} \label{approx}
So far, we have considered reward functions that are $1$-admissible.
In this section, we discuss some non-submodular function classes that can be shown to be $\alpha$-admissible for some $\alpha\geq 1.$ First, we introduce the notion of a \emph{dictator} element.  
\begin{definition}[$m$-dictator] \label{dict-def}
For some $m >0,$ assume that there exists an element $i^* \in [N]$ s.t. $f(U) \geq m, \forall U \ni i^*.$  Then we call the element $i^*$ to be an $m$-\emph{dictator}. 
\end{definition}
If the reward function $f$ is monotone, the condition $\max_{i} f(\{i\}) \geq m$ is sufficient for the existence of an $m$-dictator. The set covering interpretation of the $\alpha$-\textsc{Core}, given in Section \ref{characterization}, readily yields the following sufficient condition for the $\alpha$-admissibility of a reward function
in terms of the existence of a dictator element.

\begin{proposition} \label{dictator-prop}
Any reward function, taking values in the interval $[0,M]$ and containing an $m$-dictator, is $\frac{M}{m}$ admissible.
\end{proposition}
	\begin{proof}
	The proof follows immediately from the dual characterization given in Theorem \ref{ex-bs}. If the element $i^*$ is an $m$-dictator, we have:

	\begin{eqnarray*}
		\frac{M}{m}\sum_{S\subseteq [N]} \delta_S f(S) \geq \frac{M}{m}\sum_{U: i^* \in U} \delta_U f(U) \stackrel{(a)}{\geq} \frac{M}{m}\sum_{U: i^* \in U}m\delta_U = M \sum_{U: i^* \in U}\delta_U \stackrel{(b)}{=} M \geq f([N]), 
	\end{eqnarray*}
	where (a) follows from the fact that the element $i^*$ is an $m$-dictator and (b) follows from the fact that $\bm{\delta}$ covers the element $i^*.$
\end{proof}
In Section \ref{mdict-const} of the Appendix, we show how to efficiently construct an $M/m$-admissible vector when the reward function contains an $m$-dictator. Note that although the bound given by Proposition \ref{dictator-prop} might be loose, it is widely applicable as it does not assume any particular structure of the reward function. In the following, we apply Proposition \ref{dictator-prop} to the problem of sparse linear regression. As shown in \cite[Remark 1]{altschuler2016greedy}, the reward function $f(\cdot)$ defined below is \emph{not} submodular or even sub-additive. The following section gives an efficient construction of an admissible vector for $m$-dictatorial functions. 

\subsubsection{Efficient Construction of an admissible vector for $m$-dictatorial reward functions} \label{mdict-const}
Note that any function, which contains an $m$-dictator is called an $m$-dictatorial function. We begin our construction with the following definition. 
\begin{definition}[$m$-dictator set]
For some $m>0$, let $D_f^m\subseteq[N]$ be the set of elements such that for any set $S$ if $S\cap D_f^m\neq\emptyset$ then $f(S)\geq m$. Then, we call $D_f^m$ the $m$-dictator set associated with the set function $f$.
\end{definition}

\begin{proposition}\label{g-construct}
Assume that the reward function $f$ is monotone non-decreasing and normalized  i.e.,  $S\subseteq T \implies  f(S)\leq f(T)$, and $f(\emptyset)=0.$ Define $M \equiv f([N])$. Hence, $0 \leq f(S) \leq M, \forall S \subseteq [N].$ Suppose $f$ has a non-empty $m$-dictator set $D^m_f$ associated with it. 
\begin{itemize}
    \item [(1)]Let the set of vectors $\mathcal{G}(D_f^m, [N])$ be defined as follows:
    \begin{align}
       \mathcal{G}(D_f^m, [N]) = \left\{\bm x\in\mathbb{R}^N_{\geq0}:x_i=0, \forall  i\notin D_f^m, \textrm{ and } \sum_{i\in D_f^m}x_i=M\right\} 
    \end{align}
    Then, $\mathcal{G}(D_f^m, [N])\subseteq\core{M/m}f.$ In particular, the vector $\bm{g}$ defined as $g_i = M\mathds{1}(i=i^*), i \in [N]$ is $M/m$-admissible, where $i^* \in [N]$ is an $m$-dictator.
    \item [(2)]$f$ is $\frac{M}{m}$-admissible. 
\end{itemize}
\end{proposition}

\begin{proof}
Observe that if (1) of Proposition~\ref{g-construct} holds true, (2) follows immediately by definition. Let $\bm g\in\mathcal{G}(D_f^m, [N])$. For any subset $S \subseteq [N]$, 
\begin{align}
    \sum_{i\in S} g_i &= \sum_{i\in S\cap D_f^m} g_i + \sum_{i\in S\setminus D_f^m} g_i = \sum_{i\in S\cap D_f^m} g_i,
\end{align}
where the last equality holds by definition of $\mathcal{G}(D_f^m, [N])$. If $S\cap D_f^m=\emptyset$,
\begin{align}
    \sum_{i\in S} g_i=0\leq \frac{M}{m}f(S)
\end{align}
as $f(S)\geq0$. Suppose, $S\cap D_f^m\neq\emptyset$, 
\begin{align}
    \sum_{i\in S} g_i=\sum_{i\in S\cap D_f^m} g_i\stackrel{(a)}{\leq} M \stackrel{(b)}{\leq} \frac{M}{m}f(S)
\end{align}
where (a) holds because $\bm g\geq0$ and $\sum_{i\in D_f^m}g_i=M$ and (b) holds by the definition of the $m$-dictator set. 
So,
\begin{align}
    \sum_{i\in [N]} g_i=\sum_{i\in D_f^m} g_i = M =f([N]).
\end{align}
As $\bm g$ satisfies the conditions in Definition~\ref{admissible-rew}, $\mathcal{G}(D_f^m, [N])\subseteq \core{M/m}f$. 
\end{proof}


\paragraph{Remarks:}   We would like to emphasize that the construction of the admissible vector $\bm g$ in Proposition~\ref{g-construct} is \textit{not} exhaustive, in general. Consider a simple three-player game among the players $\{1, 2, 3\}$ with the reward function defined as follows:
\begin{align}
    f(S) = \begin{cases}
    1, & \text{if $1\in S$ or $2\in S$,}\\
    0, & \text{otherwise}
    \end{cases}
\end{align}
For $m=1$ and given $f$, $M=1$ and $D_f^m=\{1, 2\}$. So,
$\bm g=(t, 1-t, 0), t\in[0,1]$ lies in the $1$-core. It is easy to verify that $\bm g$ satisfies all the conditions mentioned in Definition~\ref{admissible-rew}. The set of $2^3-1 = 7$ constraints associated with any vector $\bm{g}$ in the core of this reward function is listed below:
\begin{align}
    g_1&\leq 1\\
    g_2&\leq 1\\
    g_3&\leq 0\\
    g_1+g_2&\leq 1\\
    g_2+g_3&\leq 1\\
    g_1+g_3&\leq 1\\
    g_1+g_2+g_3&=1.
\end{align}
The set of all solutions for the above set of constraints can be verified to be $\bm g'=(t, 1-t, 0)$ for $t\in[0,1]$. Now, consider the above three-player game with a modified reward function:
\begin{align}
    f(S)=\begin{cases}
    2, & \text{if $1\in S$}\\
    1, & \text{if $1\notin S$ but $2\in S$}\\
    0, &\text{otherwise}.
    \end{cases}
\end{align}
For $m=1$ and given $f$, $M=2$ and $D_f^m=\{1, 2\}$. Note that $\bm g=(3, 0, -1)$ lies in $\core{2}f$ but $\bm g\notin \mathcal{G}(D_f^m, [N])$

\begin{example}[Sparse Linear Regression \citep{khanna2017scalable}] \label{css}
	Consider the problem of sparse linear regression:
	\begin{eqnarray} \label{slr}
		\min_{\bm{\beta}}||\bm{y}- \bm{X}\bm{\beta}||_2^2 ~~~\textrm{s.t.}~~~ ||\bm{\beta}||_0 \leq k,
	\end{eqnarray} 
where $\bm{X} \in \mathbb{R}^{n  \times d}, d \geq n$ is a given feature matrix with unit-norm columns, and $\bm{y} \in \mathbb{R}^n$ is a response vector having unit Euclidean norm. Assume that $X$ is full rank. It is well-known that for fixed support set $S$, the optimal regressor coefficient is given by $\bm{\beta}_S^*= (\bm{X}_S^T\bm{X}_S)^{-1}\bm{X}_S^T \bm{y},$ where $\bm{X}_S$ is the sub-matrix formed by choosing columns of $\bm{X}$ corresponding to the subset $S$. Hence, the Pythagorean theorem implies that Problem \ref{slr} is equivalent to maximizing the following reward function subject to the cardinality constraint on $S$:
\begin{eqnarray} \label{slr-func}
			f(S) = ||\Pi_S \bm{y}||_2^2, 
\end{eqnarray}
		where $\Pi_S \equiv \bm{X}_S (\bm{X}_S^T\bm{X}_S)^{-1}\bm{X}_S^T$ is the projection matrix that projects any vector onto the column span of the matrix $\bm{X}_S$. Clearly, $0\leq f(S) \leq 1, \forall S \subseteq [N].$ Proposition \ref{dic-splr-proof} shows that the function $f(\cdot)$ contains an $\frac{\sigma_{\min}(\bm{X})}{4d}$-dictator. Hence, by Proposition \ref{dictator-prop}, the reward function is $\frac{4d}{\sigma_{\min}(\bm{X})}$-admissible.
\end{example} 
\begin{proposition} \label{dic-splr-proof}
	The sparse regression function $f(\cdot)$, given in \eqref{slr-func}, contains a $\frac{\sigma_{\min}(\bm{X})}{4d}$-dictator. 
\end{proposition}

\begin{proof}
	From \citet[Lemma 2]{altschuler2016greedy}, we have that for any two subsets of columns $S$ and $T$ with $f(S)\geq f(T),$ there exists a column $v^*\in S$ such that 
	\begin{eqnarray} \label{dict}
		f(T \cup v^*)-f(T)  \geq \sigma_{\min}(\bm{X}_S)\frac{(f(S)-f(T))^2}{4|S|f(S)}.
	\end{eqnarray}
	We use the above bound with $S=[d]$ and  $T =\emptyset.$ The facts that $\bm{X}$ is full rank and $|| \bm{y}||_2=1,$ imply that  $f([d])=1$. Since $f(\emptyset) =0,$ we obtain the following bound from \eqref{dict}:
	\begin{eqnarray*}
		f(\{v^*\}) \geq \frac{\sigma_{\min}(\bm{X})}{4d}. 
	\end{eqnarray*}
	
Since the reward function $f(\cdot)$ is monotone, the above inequality proves the result.
\end{proof}

\subsection{$\rho$-submodularity} \label{rho-submod}

An important class of approximately submodular functions is the class of $\rho$-submodular functions \citep{el2018combinatorial, sakaue2019greedy}. For these functions, the diminishing return property of submodular functions \eqref{diminishing} holds approximately up to a factor of $\rho>0$. Formally, a $\rho$-submodular function is defined as follows.
\begin{definition}[$\rho$-submodularity \citep{el2018combinatorial}]
	A function $f: 2^{[N]}\to \mathbb{R}$ is $\rho$-submodular iff $\exists \rho \in (0,1]$ s.t., $\forall B \subseteq [N], A \subseteq B, i \in B^c:$
\begin{eqnarray*}
\rho (f(B\cup \{i\}) - f(B)) \leq f(A\cup \{i\})- f(A). 	
\end{eqnarray*}
\end{definition}
Clearly, a $\rho$-submodular function $f$ is submodular if and only if $\rho=1.$
The paper by \citet[Proposition 2]{el2018combinatorial} gives a necessary and sufficient condition for a finite-valued monotone function to be $\rho$-submodular. In particular, their result implies that any finite-valued \emph{strictly} monotone function is $\rho$-submodular for some $\rho>0$. The following theorem gives a sufficient condition for the admissibility of $\rho$-submodular functions.
\begin{theorem} \label{approx-submod-core}
	A $\rho$-submodular function $f$ is $\nicefrac{1}{\rho}$-admissible. Furthermore, the marginal utility vector $\bm{\Delta}^\pi,$ defined in Theorem \ref{submod-core}, is $\nicefrac{1}{\rho}$-admissible. Note that the vector $\bm{\Delta}^\pi$ can be efficiently computed without knowing the value of the parameter $\rho$. 
\end{theorem}
\begin{proof}    
Let $\bm{x}$ be the marginal utility vector $\bm{\Delta}^\pi$ corresponding to some permutation $\pi.$ Relabelling the elements as $\pi(i)\gets i, \forall i \in [N],$ the components of the vector $\bm{x}$ are given as follows:  
        \begin{align}
        \begin{array}{ll}
             x_1 & = f(\{1\}),\\
             x_2 & = f(\{1,2\}) - f(\{1\}),\\
            \ & \vdots\\
             x_n & = f(\{1,2,\cdots, n\}) - f(\{1,2,\cdots, n-1\}).
        \end{array}
        \end{align}
        Then it immediately follows that $\sum_{i\in N}x_i = f(N)$. Now, consider any set $S=\{i_1,i_2,\cdots, i_k\}$ of size $k$ such that $1\le i_1<i_2<\cdots <i_k\le n$. Then, we have,
        \begin{align}
            \sum_{j\in S}x_j & = \sum_{l=1}^k x_{i_l}= \sum_{l=1}^k \big(f(\{1,2,\cdots, i_l\}) - f(\{1,2,\cdots, i_{l}-1\})\big).
        \end{align}
Using the definition of $\rho$-submodularity,         
        we obtain
        \begin{align}
            f(\{1,2,\cdots, i_l\}) - f(\{1,2,\cdots, i_{l}-1\}) & \le \rho^{-1}\big(f(\{i_1,i_2,\cdots, i_l\}) - f(\{i_1,i_2,\cdots, i_{l-1}\})\big).
        \end{align}
        Therefore,
        \begin{align}
            \sum_{j\in S}x_j & = \sum_{l=1}^k x_{i_l}\nonumber\\
            \ & \le \sum_{l=1}^k \rho^{-1}\big(f(\{i_1,i_2,\cdots, i_l\}) - f(\{i_1,i_2,\cdots, i_{l-1}\})\big)\nonumber\\
            \ & = \rho^{-1}f(S).
        \end{align}
        This shows that $\bm{x} \in \core{\rho^{-1}}{f}.$
   \end{proof}

Theorem \ref{approx-submod-core} implies that for the class of $\rho$-submodular functions, the \onscore policy can be run without knowing the value of the parameter $\rho.$ Hence, we can automatically ensure the optimal augmented regret-bound \eqref{regret-guarantee1} with the largest possible value of $\rho$ for the given sequence of approximately submodular reward functions. The following example shows an application of $\rho$-submodularity in a Bayesian Optimization setting.

\begin{example}[Batch Bayesian Optimization (\cite{halabi2020optimal})] \label{batchbayes}
    Consider the Bayesian optimization problem in a batch setting where the goal is to optimize an unknown expensive-to-evaluate function $f$ with a small number of noisy evaluations. Consider the following additive noisy observation model:
    \begin{align}
        \label{eq:noisy-batch-Bayesian}
        y = f(\bm{x}) + \epsilon,
    \end{align}
    where the unknown function $f$ is modeled using a Gaussian process (GP) with kernel $k(\bm{x},\bm{x}')$ and $\epsilon\stackrel{\textrm{i.i.d.}}{\sim}\mathcal{N}(0,\sigma^2)$. Given a set of potential optimizers $\mathcal{X}=\{\bm{x}_1,\cdots,\bm{x}_N\}$, and $S\subseteq [N]$, the posterior distribution of $f$, given observations $\bm{y}_S$ on the selected points $\{\bm{x}_i, i \in S \}$ is again a Gaussian process with variance $\sigma_S(\bm{x})^2 = k(\bm{x},\bm{x}) - k(\bm{x})^\top (\bm{K}_S+\sigma^2\bm{I})^{-1}k(\bm{x})$, where $k(\bm{x})=[k(\bm{x}_i,\bm{x})]_{i\in S}$ and $\bm{K}_S=[k(\bm{x}_i,\bm{x}_j)]_{i,j\in S}$ are the corresponding sub-matrices of the kernel matrix $\bm{K}.$ The \emph{variance reduction function}, a widely-used acquisition function, is defined as follows \citep{bogunovic2016truncated}:
    \begin{align}
        \label{eq:variance-reduction-function}
        G(S) & = \sum_{i\in [N]}(k(\bm{x}_i,\bm{x}_i)-\sigma_S^2(\bm{x}_i)).
    \end{align}
 \citet{halabi2020optimal} showed that the set function $G$ is $\rho$-submodular s.t.
    \begin{align}
        \rho & \ge \left(\frac{\lambda_{\min}(\bm{K})}{\lambda_{\min}(\bm{K})+\sigma^2}\right)^2\frac{\lambda_{\min}(\bm{K})}{\lambda_{\max}(\bm{K})}:=\eta,
    \end{align}
    where $\lambda_{\max}(\bm{K})$ and $\lambda_{\min}(\bm{K})$ are the maximum and minimum eigenvalues of $\bm{K}$, respectively. Hence, it follows from Theorem~\ref{approx-submod-core} that $G$ is $\eta^{-1}$ admissible. Therefore, the \onscore policy can learn a sequence of unknown functions modeled as a GP with time-varying kernel matrices. 
\end{example}

  \section{Proof of Theorem \ref{monotone-reward-lb}} \label{monotone-reward-lb-proof} 
  We construct a $T \times N$ random matrix $\bm{r}$ with independent rows such that each row contains exactly one $1$ corresponding to a column chosen uniformly at random from the set of all $N$ columns, and the rest of the $N-1$ entries in that row are zero. To put it formally, let $\{I_t\}_{t=1}^T$ be a sequence of i.i.d. random variables s.t. $I_t \stackrel{\textrm{i.i.d.}}{\sim} \textrm{Unif}([N]), \forall t.$ We define the $(t,i)$\textsuperscript{th} entry of the matrix to be
  \begin{eqnarray*}
  	r_{t,i} = \mathds{1}(i=I_t)~, \forall t,i.
  \end{eqnarray*}
  	Now consider a random ensemble of a sequence of linear reward functions $\{f_t\}_{t=1}^T$ defined as $f_t(S) = \sum_{i \in S}r_{ti}, \forall S \subseteq [N].$ Clearly, by construction $f_t([N])=1, \forall t \geq 1.$ Now recall that any online subset selection policy chooses  (perhaps randomly) $k$ elements out of $N$-elements at every round. Thus the r.v. $f_t(S_t)$ is Bernoulli with $\mathbb{E}(f_t(S_t)) = k/N.$ 
  	Hence, using the linearity of expectation, the expected value of the augmented regret over the random ensemble can be computed as:
  	\begin{eqnarray*}
  		\mathbb{E}\bigg[\frac{k}{N}\sum_{t=1}^T f_t([N]) - \sum_{t=1}^T f_t(S_t) \bigg] = \frac{kT}{N}- \frac{kT}{N} =0.
  	\end{eqnarray*}
  	This implies that for every online subset selection policy $\pi$, there exists a sequence of linear reward functions $\{f_t\}_{t=1}^T$ such that the augmented regret achieved by the policy $\pi$ is non-negative.

  	\section{Proof of Lemma \ref{supermodularHintBound}} \label{supermodularHintBoundproof}
    Define the set $S := \{i\in[N]: \vec{f}_i\ge \vec{h}_i\}$. Also, for any vector $\bm{v},$ define the shorthand notation $\bm{v}(S) \equiv \sum_{i \in S} v_i.$ Then,
    \begin{align}
        \lVert \vec{f} - \vec{h}\rVert_1 &= \sum_{i\in[N]} |\vec{f}_i - \vec{h}_i|\\
        &= \sum_{i\in S} (\vec{f}_i - \vec{h}_i) + \sum_{i\in S^c} (\vec{h}_i - \vec{f}_i)\\
        &= \vec{f}(S) - \vec{h}(S) + \vec{h}(S^c) - \vec{f}(S^c)\\
        &\overset{(a)}{\le} f(S) - h(S) + \vec{h}(S^c) - \vec{f}(S^c)\\
        &\overset{(b)}{=} f(S) - h(S) + h([N]) - \vec{h}(S) - \vec{f}(S^c)\\
        &\overset{(c)}{\le} f(S) - h(S) + h([N]) - \vec{h}(S) + f(S) - f([N])\\
        &\overset{(d)}{\le} f(S) - h(S) + h([N]) - h(S) + f(S) - f([N])\\
        &\le 3\max_{S\subseteq[N]}|f(S) - h(S)|
    \end{align}
    Above, in $(a)$ we have simply used the fact that $\vec{f}(S)\equiv \sum_{i \in S}f_i \le f(S)$ (as $\bm{f} \in \mathcal{B}_f$) and $\vec{h}(S)= h(S)$, since the function $h(\cdot)$ is modular. In $(b)$, we have used the modularity of the function $h$ to conclude that $\vec{h}(S^c) + \vec{h}(S)  = h([N])$ . In $(c)$, we have used the fact that $\bm{f} \in \mathcal{B}_f$ to conclude $\vec{f}(S^c) + f(S)\ge \vec{f}(S^c) + \vec{f}(S) = f([N])$. Finally, in $(d)$, we have again used the modular property of the function $h(\cdot)$. This proves the claim.

    \section{Proof of Theorem \ref{oftrlBound}} \label{OptimisticBdProof}
    
    As in the proof of Theorem \ref{main-thm}, it is enough to show the following static regret bound.
    \begin{align} \label{optimistic-bd}
        \max_{\vec{p}^*\in\Delta^N_k}\sum_{t = 1}^T \langle\vec{f}_t, \vec{p}^*\rangle - \sum_{t = 1}^T\langle \vec{f}_t, \vec{p}_t\rangle\le 4k\sqrt{\sum_{t = 1}^T ||\vec{f}_t - \vec{h}_t||_2^2}.
    \end{align}
    Then, the statement of the theorem follows after bounding the $\ell_2$ norm by $\ell_1$ and then invoking Lemma \ref{supermodularHintBound}.
    
    Now we proceed to establish the bound \eqref{optimistic-bd}.   For convenience, let us denote the LHS of the above inequality by $R_T$ and define the shorthand notation for summation $v_{1:t} \equiv \sum_{\tau=1}^t v_\tau$ where the components of $v$ are either vectors or functions. Our proof relies on Theorem 1 of \citet{pmlr-v51-mohri16}; we will first show that our algorithm satisfies all of the conditions for this theorem to be applied. In the context of this theorem, our cost functions for each $t\ge 1$ will be the linear functions $\langle \vec{f}_t, \cdot\rangle$ (and hence the gradients of the cost functions will just be the vectors $\vec{f}_t$).
    
    Recall from \cite{pmlr-v51-mohri16} that for $t\ge 1$, the regularizer $r_t$ is said to be \textit{proximal} if $    \mathrm{argmin}_{\vec{p}\in\Delta^N_k}r_t(\vec{p}) = \vec{p}_t$; it is clear from our definition of $r_t$ (line 12 of Algorithm \ref{oftrlHints})  that the regularizer $r_t$ is non-negative and proximal for $t\ge 1$. Theorem 1 of \cite{pmlr-v51-mohri16} also requires a non-negative regularization function $r_0\ge 0$; we simply define it to be the zero function, i.e $r_0(\vec{x}) = 0$ for all inputs $\vec{x}$. So, it clearly follows that $r_{0:t} = r_{1:t}$ for all $t\ge 1$.
    
    Next, note that for each $t\ge 1$ the regularizer $r_t$ is $\sigma_t$-strongly convex (with respect to the $\ell_2$ norm). This implies that $r_{0:t} = r_{1:t}$ is $\sigma_{1:t}$ strongly convex (with respect to the $\ell_2$ norm). For each $t \ge 1$, consider the norm $||\cdot||_{(t)} := \sqrt{\sigma_{1:t}}||\cdot||_2$. So, $r_{0:t}$ is $1$-strongly convex with respect to the norm $||\cdot||_{(t)}$. 

    Finally, note that the choice of $\vec{p}_t$ (line 6 of Algorithm \ref{oftrlHints}) for all $t\ge 1$ can be written as
    \begin{align}
        \vec{p}_t &= \mathrm{argmax}_{\vec{p}\in\Delta^N_k}[-r_{0:t - 1}(\vec{p}) + \langle\vec{p}, \vec{\Theta}_{t - 1} + \vec{h_t}\rangle]\\ 
        &= \mathrm{argmin}_{\vec{p}\in\Delta^N_k}[r_{0:t - 1}(\vec{p}) + \langle \vec{p}, -\vec{\Theta}_{t - 1} - \vec{h}_t\rangle\\
        &= \mathrm{argmin}_{\vec{p}\in\Delta^N_k}[r_{0:t - 1}(\vec{p}) + \langle \vec{p}, -\vec{f}_{1:t - 1}\rangle + \langle \vec{p}, -\vec{h}_t\rangle],
    \end{align}
    and this is exactly how the next point $\vec{p}_t$ is chosen in the algorithm mentioned in Theorem 1 of \cite{pmlr-v51-mohri16}. In other words, our algorithm satisfies all the conditions required for this theorem to be applied. So, applying Theorem 1 of \citet{pmlr-v51-mohri16}, we can bound the static regret as follows.
    \begin{align}
        R_T&\le r_{0:T}(\vec{p^*}) + \sum_{t = 1}^T||\vec{f}_t - \vec{h}_t||^2_{(t), *}\\
        &= r_{1:T}(\vec{p^*}) + \sum_{t = 1}^T||\vec{f}_t - \vec{h}_t||^2_{(t), *}.
    \end{align}
    Above, $||\cdot||_{(t), *}$ is the dual norm of the norm $||\cdot||_{(t)}$, i,.e 
    \begin{align}
        ||\cdot||_{(t), *} = \dfrac{1}{\sqrt{\sigma_{1:t}}}||\cdot||_2.
    \end{align}
    So, the last inequality can be written as
    \begin{align}
        R_T\le r_{1:T}(\vec{p^*}) + \sum_{t = 1}^T\dfrac{||\vec{f}_t - \vec{h}_t||_2^2}{\sigma_{1:t}}.
    \end{align}
    Now, note that the set $\Delta^N_k$ lies in the ball of radius $k$ centered at the origin, and hence the diameter of this set can be bounded above by $2k$. Hence, we can bound the regularizers as
    \begin{align}
        r_{1:T}(\vec{p}^*)\le \dfrac{\sigma_{1:T}}{2}\cdot 4k^2 = 2k^2\sigma_{1:T}, 
    \end{align}
    Hence, we have that
    \begin{align}
        R_T&\le 2k^2\sigma_{1:T} + \sum_{t = 1}^T\dfrac{||\vec{f}_t - \vec{h}_t||_2^2}{\sigma_{1:t}}\\
        &= 2k^2\sigma_{1:T} + \sum_{t = 1}^T \dfrac{\delta_t}{\sigma_{1:t}},
    \end{align}
    where in the last step, we just used the definition of $\delta_t$ (line \ref{delta-def} of Algorithm \ref{oftrlHints}). Now, by the definition of $\sigma_{t}$ (line \ref{sigma-def} of Algorithm \ref{oftrlHints}), we see that the sum $\sigma_{1:t}$ telescopes to $\sigma\sqrt{\delta_{1:t}}$, and hence we get
    \begin{align}
        R_T\le 2k^2\sigma\sqrt{\delta_{1:T}} + \sum_{t = 1}^T \dfrac{\delta_t}{\sigma\sqrt{\delta_{1:t}}}.
    \end{align}
    Now, note that for each $t\ge 1$, we have
    \begin{align}
        \dfrac{\delta_t}{\sqrt{\delta_{1:t}}} &\le \dfrac{2\delta_t}{\sqrt{\delta_{1:t}} + \sqrt{\delta_{1:t - 1}}} = 2(\sqrt{\delta_{1:t}} - \sqrt{\delta_{1:t - 1}}).
    \end{align}
    Summing the above inequality from $t = 1$ to $T$ and telescoping, we can obtain $\sum_{t = 1}^T\frac{\delta_t}{\sqrt{\delta_{1:t}}}\le 2\sqrt{\delta_{1:T}}$. Using this, we get
    \begin{align}
        R_T\le 2k^2\sigma\sqrt{\delta_{1:T}} + \dfrac{2\sqrt{\delta_{1:T}}}{\sigma}. 
    \end{align}
    Finally, picking $\sigma = \frac{1}{k}$, we get the following augmented regret bound:
    \begin{align}
        R_T\le 4k\sqrt{\delta_{1:T}} = 4k\sqrt{\sum_{t = 1}^T||\vec{f}_t - \vec{h}_t||_2^2}.
    \end{align}
   This completes the proof. 
   \subsection{Efficient Computation of the inclusion probability vector in OFTRL} \label{p-comp}

Recall that in Algorithm~\ref{oftrlHints}, we need to efficiently compute $\vec{p}_t$ by solving the following equivalent optimization problem:
\begin{align}\label{eq:main-opt}
    \vec{p}_t=\underset{\vec{p}\in\Delta^N_k}{\mathrm{argmin}}~ F_t(\vec{p})=: \underset{\vec{p}\in\Delta^N_k}{\mathrm{argmin}}\left[\sum_{\tau=1}^{t-1}\frac{\sigma_\tau}{2}\norm{\vec{p}-\vec{p}_\tau}^2 - \left\langle \vec{p}, \vec{\Theta}_{t - 1} + \vec{h}_t\right\rangle\right].
\end{align}

\paragraph{Exact computation of the inclusion probabilities $\vec{p}_t$:}

Note that 
\begin{align}
   &\underset{\vec{p}\in\Delta^N_k}{\mathrm{argmin}}\left[\sum_{\tau=1}^{t-1}\frac{\sigma_\tau}{2}\norm{\vec{p}-\vec{p}_\tau}^2 - \left\langle \vec{p}, \vec{\Theta}_{t - 1} + \vec{h}_t\right\rangle\right] \label{iter-obj}\\
   &=\underset{\vec{p}\in\Delta^N_k}{\mathrm{argmin}}\left[\sum_{\tau=1}^{t-1}\frac{\sigma_\tau}{2}\left(\norm{\vec{p}}^2 - 2\langle \vec{p}, \vec{p}_\tau\rangle\right) -\left\langle \vec{p}, \vec{\Theta}_{t - 1} + \vec{h}_t\right\rangle\right] \\
   &=\underset{\vec{p}\in\Delta^N_k}{\mathrm{argmin}}\norm{\vec{p}-\frac{\vec{\Theta}_{t - 1} + \vec{h}_t+\sum_{\tau=1}^{t-1}\sigma_\tau\vec{p}_\tau}{\sum_{\tau=1}^{t-1}\sigma_\tau}}^2.
\end{align}
Hence, the problem is equivalent to computing the Euclidean projection onto the capped simplex $\Delta^N_k$. \cite{wang2015projection} proposed an algorithm for the same with time complexity of $O(N^2)$.

\paragraph{Approximate computation of $\bm{p}_t$:}

In this section, we present a method to approximately compute $\vec{p}_t$ at each round in almost linear time, and thus overcome the $O(N^2)$ time complexity required for exact computation. We further show that, even using the approximate value of $\vec{p}_t$, we can retain the $O(\sqrt{T})$ augmented regret bound for the optimistic learning setting. To this end, we use the away steps Frank-Wolfe~(AFW) algorithm proposed by \cite{fwjulien2015} that ensures linear convergence for strongly convex functions. Formally, for a strongly convex function $F$, define $E_k = F(x_k)-F(x^*)$, where $x_k$ is the output of the AFW algorithm after $k$ iterations and $x^*$ is the optimal value. Then,
\begin{align}\label{afw-conv}
    E_{k+1}\leq (1-\theta)E_k, &&\theta\in(0,1).
\end{align}

For the sake of completeness, we present the AFW algorithm in Algorithm~\ref{afw}. In Algorithm~\ref{afw} and \ref{lmo}, $\mathcal{A}$ is the set of extreme points of $\Delta^N_k$, corresponding to the set of all $k$-sets of $[N]$.

\begin{algorithm}
    \caption{Away-steps Frank-Wolfe algorithm for computing $\vec{p}_t$}\label{afw}
    \begin{algorithmic}[1]
        \STATE \textbf{Input:} $\vec{x}^{(0)}\in\mathcal{A}$, $\varepsilon=\dfrac{\sigma\sqrt{\delta_{\tau_0}}}{200G^2T^2}$.
        \STATE Let $\vec{x}^{(0)}\in\mathcal{A}$ and $\mathcal{S}^{(0)} := \{\vec{x}^{(0)}\}$, $\alpha_{\vec{v}}^{(0)} = 1$ for $\vec{v} = \vec{x}^{(0)}$, and $0$ otherwise.

        \FOR{$i = 0$ \textbf{to} $O(\log T)$}
            \STATE Set $\vec{\nabla}_{i} := \sum_{\tau = 1}^{t - 1}\sigma_t(\vec{x}^{(i)} - \vec{p}_\tau) - \vec{\Theta}_{t - 1} - \vec{h}_t$.
            \STATE Compute $\vec{s}_i := \underset{\vec{x}\in\mathcal{A}}{\mathrm{argmin}}\langle \vec{\nabla}_i, \vec{x}\rangle$ using Algorithm \ref{lmo} and set $\vec{d}_i^{\text{FW}} := \vec{s}_i - \vec{x}^{(i)}$.

            \STATE Let $\vec{v}_i\in\underset{\vec{v}\in \mathcal{S}^{(i)}}{\mathrm{argmax}}\langle \vec{\nabla}_i, \vec{v}\rangle$ and set $\vec{d}_i^{\text{A}} := \vec{x}^{(i)} - \vec{v}_i$.

            \IF{$\langle -\vec{\nabla}_i, \vec{d}_i^{\text{FW}}\rangle\le \epsilon$}
                \STATE \textbf{return} $\vec{x}^{(i)}$
            \ENDIF

            \IF{$\langle -\vec{\nabla}_i, \vec{d}_i^{\text{FW}}\rangle \ge \langle -\vec{\nabla}_i, \vec{d}_i^{\text{A}}\rangle$}
                \STATE $\vec{d}_i := \vec{d}_i^{\text{FW}}$ and $\gamma_{\text{max}} := 1$.
            \ELSE
                \STATE $\vec{d}_i := \vec{d}_i^{\text{A}}$ and $\gamma_{\text{max}} := \frac{\alpha_{\vec{v}_i}}{1 - \alpha_{\vec{v}_i}}$.
            \ENDIF

            \STATE Line-search: $\gamma_i\in\underset{\gamma\in[0, \gamma_{\text{max}}]}{\mathrm{argmin}} \sum_{\tau = 1}^{t - 1}\frac{\sigma_\tau}{2}||\vec{x}^{(i)} + \gamma\vec{d}_i - \vec{p}_\tau||_2^2 - \langle \vec{x}^{(i)} + \gamma\vec{d}_i, \vec{\Theta}_{t - 1} + \vec{h}_t\rangle$.

            \STATE Update $\vec{x}^{(i + 1)} := \vec{x}^{(i)} + \gamma_i\vec{d}_i$ and the weights $\vec{\alpha}^{(i + 1)}$ respectively.

            \STATE Update $\mathcal{S}^{(i + 1)} := \{\vec{v}\in\mathcal{A}: \alpha_{\vec{v}}^{(i + 1)} > 0\}$.
            \ENDFOR 
    \end{algorithmic}
\end{algorithm}

\begin{algorithm}
    \caption{Linear Minimization Oracle for $\mathcal{A}$}\label{lmo}
    \begin{algorithmic}[1]
        \STATE \textbf{Input}: Cost vector $\vec{c}\in\mathbb{R}^N$.
        
        \STATE \textbf{Output:} $\underset{\vec{x}\in\mathcal{A}}{\mathrm{argmin}}\langle \vec{c}, \vec{x}\rangle$.

        \STATE Sort the elements of the set $[N]$ in non-decreasing order according to the coordinates of $\vec{c}$. Let the sorted set be $\{j_1, j_2, ..., j_N\}$.

        \STATE Let $\vec{p}$ be such that $\vec{p}_{j_i} = 1$ for all $1\le i\le k$ and $\vec{p}_{j_i} = 0$ for $k < i \le N$.

        \STATE \textbf{return} $\vec{p}$.
    \end{algorithmic}
\end{algorithm}

\paragraph{Regret analysis with approximately optimal inclusion probability vector $\vec{p}_t$:}

Suppose $\Tilde{\vec{p}}_t$ be the approximate solution of $\vec{p}_t$ obtained using the AFW algorithm \ref{afw}. We have,
\begin{align}
\textrm{Regret}^{\textrm{AFW}}_T &= \max_{\bm{p}^* \in \Delta^N_k} \sum_{t \leq T} \langle {\bm{f}}_t, \bm{p}^* \rangle - \sum_{t \leq T} \langle {\bm{f}}_t, \Tilde{\bm{p}}_t \rangle\\
&=\max_{\bm{p}^* \in \Delta^N_k} \sum_{t \leq T} \langle {\bm{f}}_t, \bm{p}^* \rangle - \sum_{t \leq T} \langle {\bm{f}}_t, \bm{p}_t \rangle + \sum_{t \leq T} \langle {\bm{f}}_t, \bm{p}_t -\Tilde{\bm{p}}_t \rangle\\
&\leq \textrm{Regret}^{\textrm{exact}}_T + G\sum_{t \leq T}\norm{\bm{p}_t -\Tilde{\bm{p}}_t},\label{eq:reg-UB}
\end{align}
where in the last line we have used the Cauchy-Schwarz inequality and $\norm{{\bm{f}}_t}\leq G, \forall t$. Denote the OFTRL objective function \eqref{iter-obj} on round $t$ by $F_t(\bm{p}).$ Now, using the definitions of $\delta_\tau$ and  $\sigma_\tau$ from steps \eqref{delta-def} and \eqref{sigma-def} of Algorithm \ref{oftrlHints}, we have
\begin{align}
    \nabla^2 F_t(\vec{p}) = \left(\sum_{\tau=1}^{t-1}\sigma_\tau\right)\bold{I}\succeq \sigma\sqrt{\sum_{\tau=1}^{t-1}\delta_\tau}\bold{I}\succeq \sigma\sqrt{\delta_{\tau_0}}\bold{I},
\end{align}
where $\tau_0=\min\{\tau\in[t-1]: \delta_\tau>0\}$. Note that if $\delta_\tau=0, \forall \tau\in[t-1]$, then $F_t$ is linear, and the exact optimal value can be computed in $O(N\log N)$ time, i.e. by sorting the elements of coefficient of $\vec{p}$. Without loss of generality, assume that there exists a feasible value of $\tau_0$. Then, all objective functions $F_\tau$ are strongly convex for $\tau\geq\tau_0$. Recall that for a $\gamma$-strongly convex function $F:\mathcal{X}\to\mathbb{R}$, $ \forall x, y \in \mathcal{X}$,
\begin{align}
    F(y)-F(x)\geq \langle \nabla F(x), y-x\rangle+\frac{\gamma}{2}\norm{y-x}^2.
\end{align}
So, for all $F_\tau, \tau\geq\tau_0$, we have
\begin{align}\label{strong-convex}
   F_\tau(\Tilde{\vec{p}}_\tau)-F_\tau(\bm{p}_\tau)\geq \langle \nabla F_\tau(\bm{p}_\tau), \Tilde{\vec{p}}_t-\bm{p}_\tau\rangle+\frac{\sigma\sqrt{\delta_{\tau_0}}}{2}\norm{\Tilde{\vec{p}}_\tau-\bm{p}_\tau}^2\geq \frac{\sigma\sqrt{\delta_{\tau_0}}}{2}\norm{\Tilde{\vec{p}}_\tau-\bm{p}_\tau}^2, 
\end{align}
where the last inequality holds because $\langle \nabla F_\tau(\bm{p}_\tau), \Tilde{\vec{p}}_\tau-\bm{p}_\tau\rangle\geq0$ due to optimality conditions of smooth convex optimization over a convex constraint set. Assume that we continue the AFW algorithm for each $\tau>\tau_0$ till it reaches an $\varepsilon$-optimal solution, i.e. $F_\tau(\Tilde{\vec{p}}_\tau)-F_\tau(\bm{p}_\tau)\leq \varepsilon$. So, using the result of \eqref{strong-convex},  
\begin{align}
    \norm{\Tilde{\vec{p}}_\tau-\bm{p}_\tau} \leq \sqrt{\frac{2\varepsilon}{\sigma\sqrt{\delta_{\tau_0}}}}.
\end{align}
Substituting in \eqref{eq:reg-UB}, we have
\begin{align}
 \textrm{Regret}^{\textrm{AFW}}_T &\leq \textrm{Regret}^{\textrm{exact}}_T + G\sum_{t \leq T}\norm{\bm{p}_t -\Tilde{\bm{p}}_t} \leq    \textrm{Regret}^{\textrm{exact}}_T + GT\sqrt{\frac{2\varepsilon}{\sigma\sqrt{\delta_{\tau_0}}}}.
\end{align}
Setting $\varepsilon=\dfrac{\sigma\sqrt{\delta_{\tau_0}}}{200G^2T^2}$ makes the second term a constant, so the overall regret bound is $O(\sqrt{T})$. Suppose the AFW algorithm terminates after $K$ iterations. Using the result of \eqref{afw-conv} for upper bounding the optimality gap and substituting the value of $\varepsilon$, we have 
\begin{align}
    E_{K}\leq e^{-K\theta}E_0\leq \varepsilon \implies K\geq\frac{1}{\theta}\log\left(\frac{E_0}{\varepsilon}\right)=\frac{1}{\theta}\log\left(\frac{200E_0G^2T^2}{\sigma\sqrt{\delta_{\tau_0}}}\right)=O(\log T).
\end{align}
So, the total time complexity per round is $O(N\log N\log T)$, which is better than $O(N^2)$ complexity \citep{wang2015projection} for high-dimensional problems.

  \section{Online Subset selection in the Priced Feedback Model (Active Learning)}\label{priced}
Throughout the paper, we have considered the \emph{full-information} setting, where the reward functions are revealed (\emph{e.g.,} through an evaluation oracle) at the end of each round.
In this section, we consider a partial information setting in which the learner has to pay a fixed cost $C$ each time it chooses to access the reward function~$f_t$ (and consequently $\boldsymbol g_t$). It does not pay any cost for round $t$ if it chooses not to access the reward function $f_t$ on round $t.$ This learning model is similar to the active learning framework in the supervised learning literature \citep{streeter2007onlinereport}. 
We first aim to establish a sublinear standard regret bound for linear rewards and then extend it for non-linear reward functions and augmented regret. In this case, the (augmented) regret is the sum of the usual notion of (augmented) regret and the cost accumulated for observing the reward functions. We apply a learning policy similar to \citet{streeter2007onlinereport}, described in Algorithm~\ref{priced-linear}.

\begin{algorithm}
\caption{Online Subset Selection Policy with \textsc{Linear-rewards} (Priced feedback model)}
\label{priced-linear}
\begin{algorithmic}[1]
\REQUIRE Sequence of linear reward functions $\{f_t\}_{t \geq 1}$ with coefficients $\{\bm{g}_t\}_{t\geq 1}, \eta >0, \varepsilon\in(0,1]$
\STATE $\bm{p} \gets (\frac{k}{N}, \frac{k}{N}, \ldots, \frac{k}{N}).$
\FOR {every round $t$:}
\STATE Sample $Z_t\sim$ Bernoulli($\varepsilon$).
\IF{$Z_t=1$}
\STATE Sample a $k$-set according to the inclusion probability vector $\bm{p}$ using Algorithm~\ref{uneq}.
\STATE Pay cost $C$ and observe $\boldsymbol g_t$. Set $\bm{\hat{g}}_t=\boldsymbol g_t/\varepsilon$.
\ELSE
\STATE Set $\bm{\hat{g}}_t=\boldsymbol 0$.
\ENDIF
\STATE Feed the estimated vector $\bm{\hat{g}}_t$ and update the inclusion probability vector $\bm{p}$ according to the \textsc{FTRL}($\eta$) update \eqref{ftrl-opt}.
\ENDFOR
\end{algorithmic}
\end{algorithm}

\subsection{Regret Analysis for Linear Rewards} Let $Z_1, Z_2,\dots, Z_T$ be i.i.d. Bernoulli random variables with parameter $\varepsilon$. So, the reward vector at any timestep $t$ is given by:
\begin{align}
    \bm{\hat{g}}_t = \frac{Z_t\bm{g}_t}{\varepsilon}\implies \mathbb{E}[\bm{\hat{g}}_t] = \bm{g}_t
\end{align}
As described earlier, the regret for the priced feedback model~(PFM) is given by:
\begin{eqnarray}\label{eq:pfm}
    \textrm{Regret}^{\textrm{PFM}}_T = \textrm{Regret}^{\textrm{FTRL}}_T + \textrm{(\# of times cost $C$ is paid)}
\end{eqnarray}

 Now recall the following standard static regret bound for the \textsc{FTRL}($\eta$) policy from \citet[Theorem 5.2]{hazan2019introduction}. For a bounded, convex and closed set $\Omega$ and a strongly convex regularization function $R: \Omega \to \mathbb{R},$ consider the standard \textsc{FTRL} updates, \emph{i.e.},
 \begin{eqnarray}\label{ftrl-updates}
  \bm{x}_{t+1} = \arg\max_{\bm{x} \in \Omega}\left[ \left(\sum_{s=1}^t \nabla_s^\top\right) \bm{x} - \frac{1}{\eta}R(x), \right]   
 \end{eqnarray} 
 where $\nabla_s= \nabla f_t(\bm{x}_s), \forall s$. Taking $\Omega \equiv \Delta^N_k$ to be the capped-simplex defined by \eqref{necsuf1} and \eqref{necsuf2} and $R(\bm{x}) \equiv \sum_{i=1}^N x_i \ln x_i$ to be the entropic regularizer, the regret of the \textsc{FTRL}($\eta$) policy can be upper-bounded as follows:
  \begin{eqnarray}\label{reg-bd-main}
  \textrm{Regret}^{\textrm{FTRL}}_T \leq 2 \eta \sum_{t=1}^T || \nabla_t||_{*,t}^2 + \frac{R(\bm{u})-R(\bm{x}_1)}{\eta},
  \end{eqnarray}

 where the quantity $||\nabla_t||_{*,t}^2$ denotes the square of the dual norm of the vector induced by the Hessian of the regularizer evaluated at some point $\bm{x}_{t+1/2}$ lying in the line segment connecting the points $\bm{x}_t$ and $\bm{x}_{t+1}$. Hence, for the \textsc{Linear-reward} problem with the sequence of reward vectors $\hat{\bm g}_1, \dots, \hat{\bm g}_T$, we have:
\begin{align}
    \max_{\bm{p}^* \in \Delta^N_k} \sum_{t \leq T} \langle \hat{\bm{g}}_t, \bm{p}^* \rangle - \sum_{t \leq T} \langle \hat{\bm{g}}_t, \bm{p}_t \rangle \leq 2 \eta \sum_{t=1}^T || \nabla_t||_{*,t}^2 + \frac{R(\bm{u})-R(\bm{x}_1)}{\eta},
\end{align}
The following diagonal matrix gives the Hessian of the entropic regularizer: 
 \begin{eqnarray*}
 \nabla^2 R(\bm{p}_{t+1/2}) = \textrm{diag}([p_1^{-1}, p_2^{-1}, \ldots, p_N^{-1}])\implies || \nabla_t||_{*,t}^2 =\langle \bm p_t, \hat{\bm g}^2_t\rangle,
 \end{eqnarray*}
In the above, the vector $\hat{\bm g}_t^2$ is obtained by squaring each component of the vector $\hat{\bm g}_t.$ To bound the second term in \eqref{reg-bd-main}, define a probability distribution $\tilde{\bm{p}} = \bm{p}/k.$ We have 
 \begin{eqnarray*}
 0 \geq R(\bm{p})  = \sum_i p_i \ln p_i = - k \sum_i \tilde{p}_i \ln \frac{1}{p_i}\stackrel{\textrm{(Jensen's inequality)}}{\geq} -k \ln \sum_i \frac{\tilde{p}_i}{p_i} = -k \ln \frac{N}{k}.   	
 \end{eqnarray*}
Hence, for this problem, the FTRL regret bound in \eqref{reg-bd-main} simplifies to the following:
\begin{eqnarray}\label{eq:pivotal}
 \max_{\bm{p}^* \in \Delta^N_k} \sum_{t \leq T} \langle \hat{\bm{g}}_t, \bm{p}^* \rangle - \sum_{t \leq T} \langle \hat{\bm{g}}_t, \bm{p}_t \rangle \leq \frac{k}{\eta} \ln \frac{N}{k}+ 2\eta \sum_{t=1}^T \langle \bm p_t, \hat{\bm g}^2_t\rangle. 
\end{eqnarray}
Combining \eqref{eq:pfm} and \eqref{eq:pivotal}:
\begin{align}
 \textrm{Regret}^{\textrm{PFM}}_T &= \max_{\bm{p}^* \in \Delta^N_k}  \sum_{t \leq T} \langle \hat{\bm{g}}_t, \bm{p}^* \rangle - \sum_{t \leq T} \langle \hat{\bm{g}}_t, \bm{p}_t \rangle + \textrm{(\# of times cost $C$ is paid)}
 \\
 &\leq \frac{k}{\eta} \ln \frac{N}{k}+ 2\eta \sum_{t=1}^T \langle \bm p_t, \hat{\bm g}^2_t\rangle + \sum_{t=1}^T CZ_t. 
\end{align}
Observe,
\begin{align}
   \langle \bm p_t, \hat{\bm g}^2_t\rangle = \left\langle \bm p_t, \frac{Z_t\bm g^2_t}{\varepsilon^2}\right\rangle = \frac{Z_t}{\varepsilon^2}\langle \bm p_t, \bm g^2_t\rangle \leq \frac{Z_t}{\varepsilon^2}\sum_{i=1}^N g_{t,i}^2\leq\frac{Z_tG^2}{\varepsilon^2}.
\end{align}
Let $Y=\sum_{t=1}^TZ_t$. Substituting the values, we get 
\begin{eqnarray*}
\textrm{Regret}^{\textrm{PFM}}_T \leq \frac{k}{\eta} \ln \frac{N}{k}+ \frac{2\eta YG^2}{\varepsilon^2} + CY. 
\end{eqnarray*}

Taking expectations of both sides with respect to the random choice of the policy and using Jensen's inequality, we have
\begin{align}
\mathbb{E}\left[\textrm{Regret}^{\textrm{PFM}}_T\right] &= \max_{\bm{p}^* \in \Delta^N_k} \sum_{t \leq T} \langle {\bm{g}}_t, \bm{p}^* \rangle - \sum_{t \leq T} \langle {\bm{g}}_t, \bm{p}_t \rangle + \mathbb{E}\left[\textrm{(\# of times cost $C$ is paid)}\right]\\
&\leq \frac{k}{\eta} \ln \frac{N}{k}+ \frac{2\eta TG^2}{\varepsilon} + \varepsilon CT,  
\end{align}
where we have used the fact that the estimated rewards are unbiased and $\mathbb{E}[Y]=\varepsilon T$. Setting $\eta=\sqrt{\frac{\varepsilon k\ln (N/k)}{2TG^2}}$,
\begin{align}
\mathbb{E}\left[\textrm{Regret}^{\textrm{PFM}}_T\right] \leq \sqrt{\frac{8G^2Tk\ln{N/k}}{\varepsilon}} + \varepsilon CT.  
\end{align} Setting $\varepsilon = \left(\dfrac{2G^2k\ln{ (N/k)}}{TC^2}\right)^{1/3}$, we have
\begin{align}\label{linear-reward-priced}
\mathbb{E}\left[\textrm{Regret}^{\textrm{PFM}}_T\right] \leq 4G^{2/3}\left(k\ln\frac{N}{k}\right)^{1/3}T^{2/3}=O\left(T^{2/3}\right),  
\end{align}
Note that for sufficiently large $T$, we have $\varepsilon\in(0,1]$.

\subsection{Outline of Augmented Regret analysis for non-linear reward functions}

In the non-linear reward setting, we define the subset selection problem for $\alpha$-admissible functions in the priced feedback model as follows: at any round $t$, sample $Z_t\sim$ Bernoulli($\varepsilon$). If $Z_t=1$, the learner plays a $k$-set $S_t$. The adversary reveals the reward function $f_t$ and the learner computes $\bm g_t\in\core{\alpha}{f_t}$. Then, we use Algorithm~\ref{priced-linear} to construct an unbiased estimator $\hat{\bm g}_t$ and subsequently update the weights. Finally, the same augmented regret bound given in \eqref{regret-guarantee1} follows immediately in the priced feedback setting using the inequalities \eqref{est1} and \eqref{est2} and the priced feedback regret bound in \eqref{linear-reward-priced}. From Proposition \ref{core-bd}, for $\alpha$-admissible reward functions taking values in $[0, M],$ we can take $G \equiv \alpha M\sqrt{2}$. It is easy to see that the space and time complexity of this version is of the same order as in the complete information setting.

\appendix

\end{document}